\documentclass[11pt,a4paper,logo]{googledeepmind}

\setleftlogo[110pt]{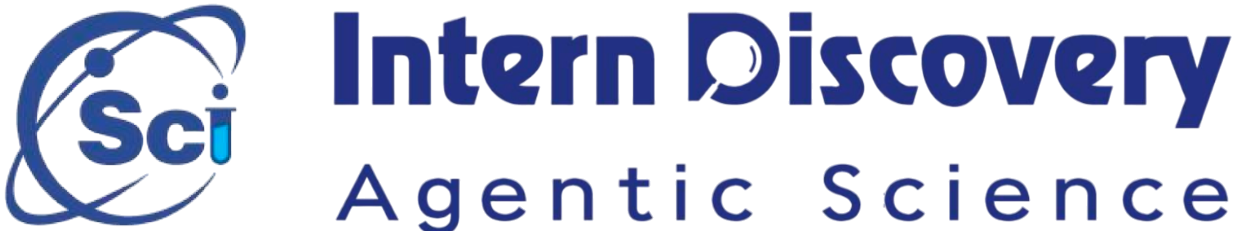}  
\setrightlogo[160pt]{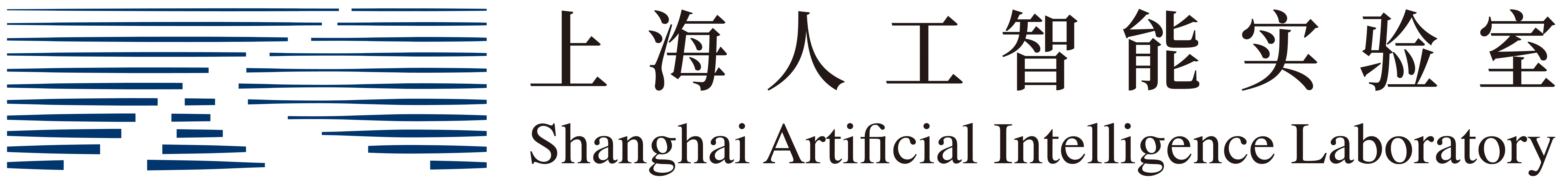}

\usepackage[T1]{fontenc}
\usepackage{pifont}

\usepackage[
    natbib=true,
    backend=biber,
    style=numeric, 
    sorting=none 
]{biblatex}

\AtEveryBibitem{\clearfield{month}}
\AtEveryBibitem{\clearfield{day}}
\usepackage{csquotes}
\usepackage[table,xcdraw,dvipsnames]{xcolor}
\usepackage[many,breakable]{tcolorbox}

\newcommand{\ProjectName}{SciEvalKit}

\title{\logo SciEvalKit: An Open-source Evaluation Toolkit for Scientific General Intelligence}

\newcommand{\homepage}{\raisebox{-1.5pt}{\includegraphics[height=1em]{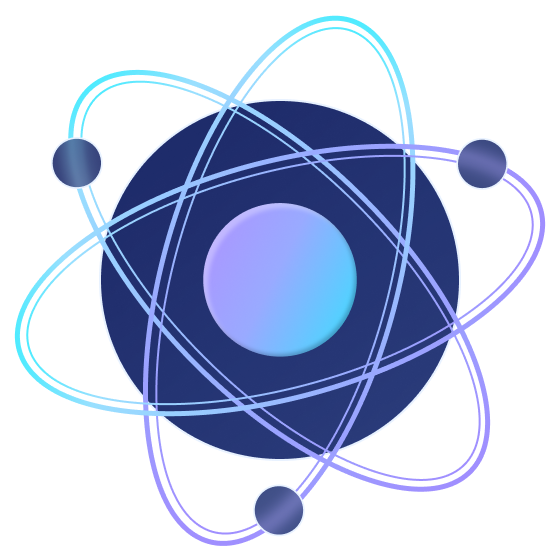}}}
\newcommand{\github}{\raisebox{-1.5pt}{\includegraphics[height=1em]{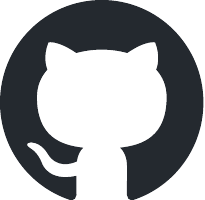}}}

\newcommand{\logo}{\raisebox{-1.5pt}{\includegraphics[height=1em]{imgs/opencompass.png}}}

\author{Shanghai Artificial Intelligence Laboratory and Community Contributors\footnote{SciEvalKit contributors can join the author list of the report based on their contribution to the repository. Specifically, it requires 3 major contributions (implement a new benchmark, foundation model, or contribute a major feature). We will update the report quarterly and an additional section that details each developer’s contribution will be appended in the next update.}
}

\usepackage{pdflscape}

\usepackage{textcomp}
\usepackage{rotating}

\usepackage{setspace}
\usepackage{microtype} 

\usepackage{graphicx}
\usepackage{subcaption}
\usepackage{caption}

\usepackage{booktabs}
\usepackage[table]{xcolor}

\usepackage{array}
\usepackage{threeparttable}
\usepackage{multirow}

\usepackage{amsmath}
\usepackage{siunitx}

\usepackage{enumitem}
\usepackage{float}
\usepackage{seqsplit}
\usepackage{framed}

\usepackage{tikz}

\usepackage{ragged2e}

\usepackage{makecell}[most]
\usepackage{tabularray}    
\usepackage{longtable,booktabs,multirow,array,hyperref}

\usepackage[symbol]{footmisc}
\newcolumntype{Y}{>{\RaggedRight\arraybackslash}X}

\usepackage{hyperref}
\hypersetup{
    colorlinks=true,
    linkcolor=blue, 
    citecolor=blue,  
    filecolor=black,
    urlcolor=blue    
}

\usepackage{tocloft}
\usepackage{xcolor}
\definecolor{CloseBG}{HTML}{DCEEF9}  
\definecolor{OpenBG}{HTML}{FCE8F1}  

\usepackage{graphicx}
\usepackage{adjustbox} 
\newcommand{\gold}{\raisebox{-3pt}{\includegraphics[height=12pt]{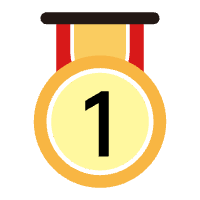}}}
\newcommand{\silver}{\raisebox{-3pt}{\includegraphics[height=12pt]{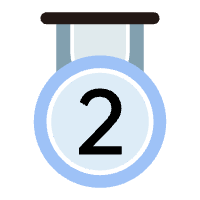}}}
\newcommand{\copper}{\raisebox{-3pt}{\includegraphics[height=12pt]{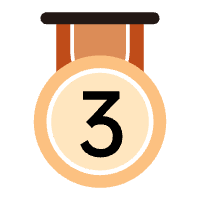}}}

\usepackage{etoolbox}
\usepackage{arydshln}
\usepackage{url}
\makeatletter
\patchcmd{\@tocline}
    {\hfil}
    {\leaders\hbox{\hfil}\hfil}
    {}{}
\makeatother

\begin{abstract}
\textbf{Abstract} \\
We introduce \ProjectName, a unified benchmarking toolkit designed to evaluate AI models for science across a broad range of scientific disciplines and task capabilities. Unlike general-purpose evaluation platforms, {\ProjectName} focuses on the core competencies of scientific intelligence, including Scientific Multimodal Perception, Scientific Multimodal Reasoning, Scientific Multimodal Understanding,  Scientific Symbolic Reasoning, Scientific Code Generation, Science Hypothesis Generation and Scientific Knowledge Understanding. It supports six major scientific domains, spanning from physics and chemistry to astronomy and materials science.
{\ProjectName} builds a foundation of expert-grade scientific benchmarks, curated from real-world, domain-specific datasets, ensuring that tasks reflect authentic scientific challenges. The toolkit features a flexible, extensible evaluation pipeline that enables batch evaluation across models and datasets, supports custom model and dataset integration, and provides transparent, reproducible, and comparable results.
By bridging capability-based evaluation and disciplinary diversity, {\ProjectName} offers a standardized yet customizable infrastructure to benchmark the next generation of scientific foundation models and intelligent agents. The toolkit is open-sourced and actively maintained to foster community-driven development and progress in AI4Science. \\

\homepage\ \textbf{Page} \texttt{\url{https://opencompass.org.cn/Intern-Discovery-Eval/rank}}

\github\ \textbf{Code} \texttt{\url{https://github.com/InternScience/SciEvalKit}}

\end{abstract}

\usepackage{listings}
\begin{document}
\sloppy 
\maketitle

\begin{figure*}[h]
  \centering
  \includegraphics[width=0.9\linewidth]{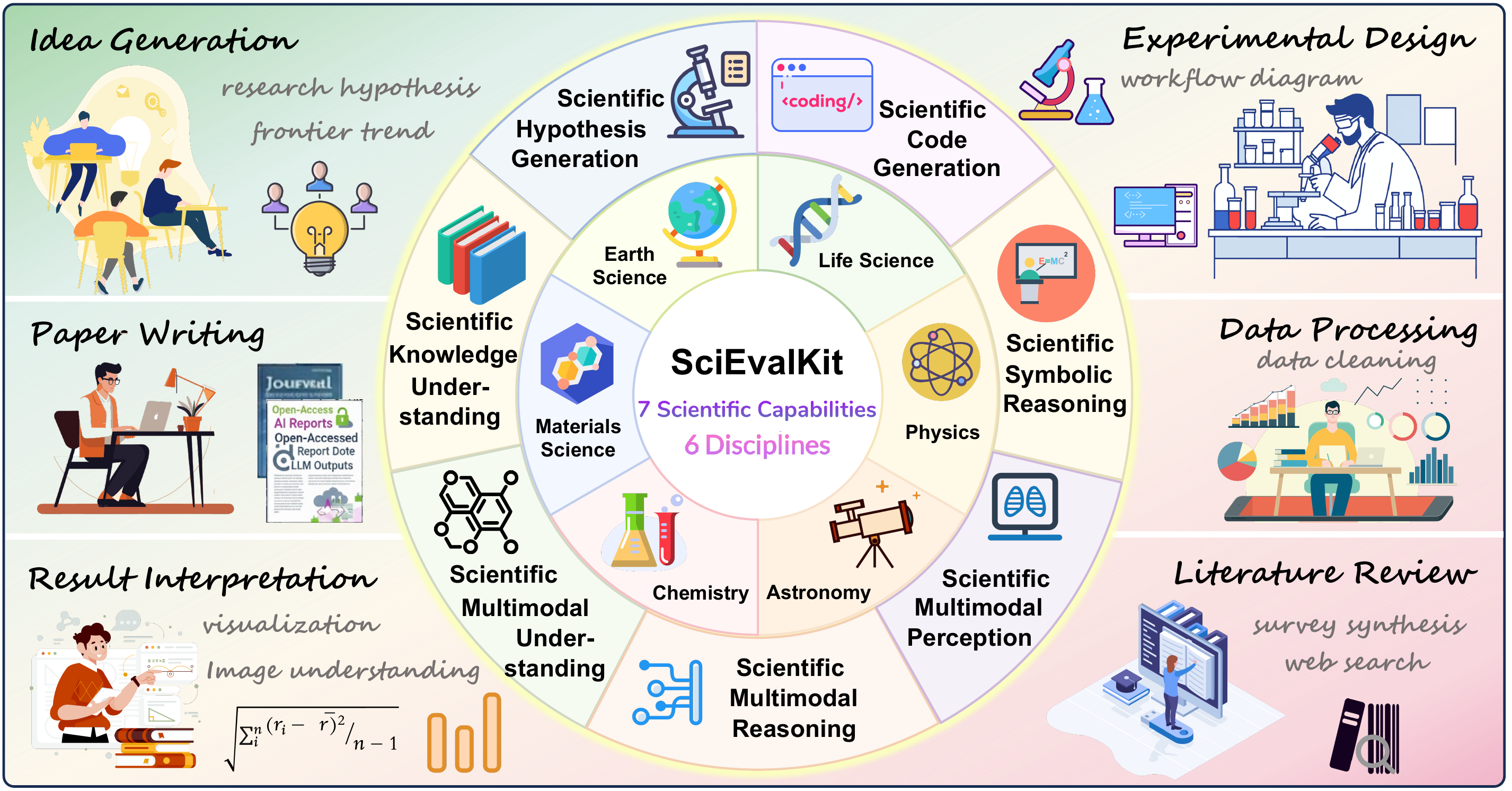}
  \caption{\centering Overview of the {\ProjectName} scientific intelligence evaluation framework.
  }
  \label{fig:overview}
\end{figure*}

\clearpage

\tableofcontents
\enlargethispage{1cm} 
\thispagestyle{fancy} 

\clearpage

\section{Introduction}


The advances in large language models (LLMs) have demonstrated remarkable general‐purpose reasoning~\cite{huang2023towards,jin2024impact,wang2024llmrg,yang2024buffer} and broad knowledge retrieval~\cite{long2024generative,wang2025astute,ren2025investigating}.
Recently, researchers are increasingly interested in probing whether these models demonstrate key facets of scientific intelligence such as conceptual understanding~\cite{storey2025large,schiappa2024probing,jin2025exploring,jiang2025ewe}, symbolic reasoning~\cite{qilarge,fang2024large,ma2024sciagent}, and hypothesis-driven exploration~\cite{koneru2024can,yang2024large,xu2025manalyzer,hu2025flowsearch}.
Despite encouraging progress on individual benchmarks~\cite{tian2024scicode,babe2024studenteval,roberts2024scifibench}, current evaluations largely focus on surface-level correctness or narrow task-specific metrics, and therefore fail to assess whether LLMs can truly operate across the full spectrum of scientific reasoning. Real-world scientific problem solving fundamentally differs from generic reasoning: it requires conceptual abstraction, symbolic manipulation, hypothesis formation, multi-step procedural thinking, and the ability to interpret structured visual representations such as chemical diagrams~\cite{ruan2024automatic,ye2025drugassist}, protein structures~\cite{li2024prosst,shen2024fine}.
Yet existing benchmarks neither capture this holistic view nor systematically evaluate these capabilities across scientific disciplines, modalities, and cognitive dimensions.

From a cognitive perspective, scientific reasoning is inherently structural, relational, and multi-representational. 
The famous DSRP~\cite{CABRERA2008299} Theory which represents Distinctions, Systems, Relationships, Perspectives respectively posits that all complex reasoning emerges from these four fundamental cognitive patterns. Moreover, it emphasizes that individuals can improve their reasoning capacities by explicitly engaging with these four elements.
This theoretical perspective provides a principled foundation for the core capabilities required to model scientific intelligence, prompting us to move beyond factual memorization or pattern recognition toward evaluating large language models' abilities in relational reasoning, structured reasoning, and representational alignment across textual, symbolic and visual modalities.
Grounded in this framework, we propose a taxonomy of seven core dimensions of scientific intelligence as shown in Fig.~\ref{fig:overview} that reflect essential capabilities in modern scientific practice: (1) \textbf{Scientific Knowledge Understanding} which assesses the models' grasp of  domain-specific concepts and factual relationships. (2) \textbf{Scientific Code Generation} which captures the ability to translate scientific descriptions and algorithmic procedures into executable code. (3) \textbf{Scientific Symbolic Reasoning} which evaluates the manipulation of equations, physical laws, symbolic expressions, and structured notation. (4) \textbf{Scientific Hypothesis Generation} which measures the capabilities to propose plausible hypotheses and research directions. (5) \textbf{Scientific Multimodal Perception} which focuses on the entity localisation and grounding in paired visual–text inputs. (6) \textbf{Scientific Multimodal Reasoning} which involves chain-of-thought inference with domain scientific image data typically found in research papers. (7) \textbf{Scientific Multimodal Understanding} which probes rigorous interpretation of raw scientific data.


Existing benchmarks typically assess isolated abilities such as factual question answering~\cite{wang2024causal,hu2024towards,louis2024interpretable}, code completion~\cite{tian2024scicode,joseph2025astrovisbench} and visual recognition~\cite{wu2024transferring,liu2024democratizing}. However, they seldom capture the reasoning processes for realistic scientific workflows~\cite{liu2025atlas,wan2025deepresearch,xu2025probing,hu2025flowsearch}. Most multimodal benchmarks are confined to generic image caption tasks~\cite{chen2024sharegpt4v,jin2024chat}, lacking expert-level semantic alignment capabilities for scientific images. Furthermore, mainstream leaderboards~\cite{park2024open,singh2024legobench} often emphasize overall scores and overlook the detailed differences between scientific dimensions.

\begin{figure*}[t]
  \centering
  \includegraphics[width=\linewidth]{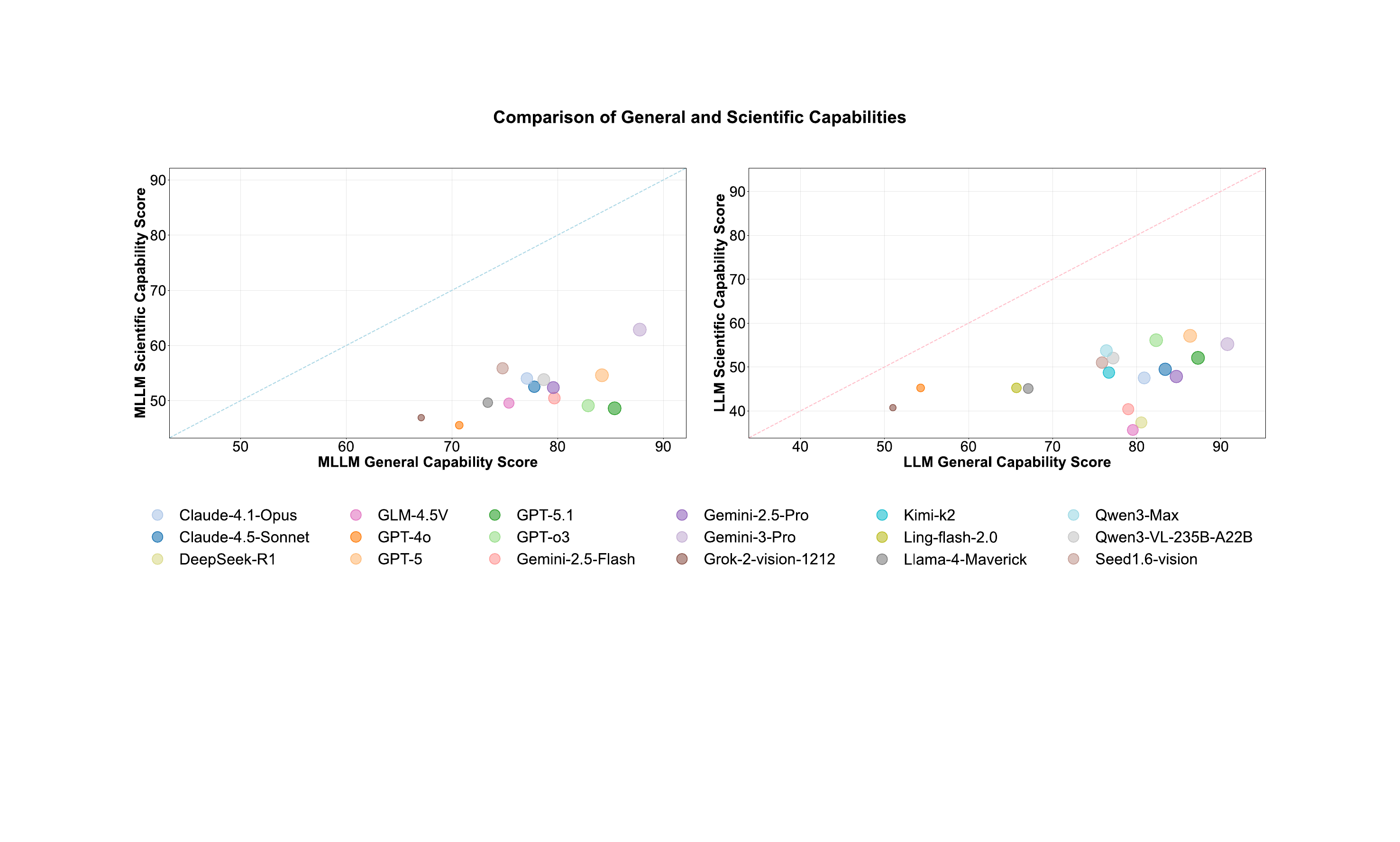}
  \caption{%
    \centering Comparison of model performance on scientific versus general tasks.
  }
  \label{fig:general_scientific_comparison}
\end{figure*}

Frontier models~\cite{zhang2025large,hu2025survey} are predominantly engineered and tuned for general-purpose utility such as handling dialogue~\cite{wang2024full,yu2024cosafe,bai2024mt}, broad-domain retrieval or generic reasoning tasks~\cite{chen2024premise,parmar2024logicbench,yang2024buffer}.  Yet science tasks impose specific requirements: precise symbolic manipulation~\cite{guo2025self}, code–execution fidelity~\cite{guo2025self}, and the ability to align dense textual arguments with highly specialised diagrams or experimental data~\cite{tang2025eigen,wang2025scireasoner,he2025radarqa,wang2025omniearth}.  To quantify whether current general models meet these standards, we compare each model's score on general benchmarks against its score on corresponding scientific domain benchmarks (shown in Fig.~\ref{fig:general_scientific_comparison}).  The strongest model Gemini-3 Pro already approaches 90 score on general tasks, but it and every other model falls below 60 score once they are probed under rigorous scientific scenario.  This systematic gap highlights the need for an integration of general and specialized capabilities that broad instruction tuning must be fused with expert-level skills in coding, symbolic reasoning and diagram understanding.

To address this critical gap, we introduce \ProjectName, an open-source evaluation toolkit and leaderboard for scientific intelligence that assesses large language models (LLMs) and multimodal large language models (MLLMs) across seven core dimensions. {\ProjectName} integrates over 15 expert-curated benchmarks spanning six major scientific disciplines including life sciences, chemistry, earth sciences, materials science, physics, and astronomy. It supports multimodal inputs and integrates both direct text-answer scoring and code-execution scoring, along with semantic LLM-as-a-judge validation and expert-aligned criteria. Through this unified framework, {\ProjectName} establishes a transparent, cognitively grounded, and scientifically credible evaluation paradigm for next-generation scientific AI systems.

Using \ProjectName, we evaluate the scientific intelligence of cutting-edge LLMs and MLLMs from both proprietary and open-source providers. Our findings reveal substantial disparities among models: while most achieve moderate-to-strong performance in knowledge understanding, capabilities such as symbolic reasoning and code generation remain underdeveloped. Notably, these shortcomings persist even among vision-enabled or instruction-tuned models, highlighting the need for capability-oriented evaluation.

In summary, this work contributes: (1) a seven-dimensional capability taxonomy grounded in expert-defined reasoning demands, (2) \ProjectName: an open-source, multimodal, execution-aware, and expert-aligned evaluation toolkit, and (3) a comprehensive benchmark analysis of leading LLMs, uncovering critical gaps in their readiness for real scientific problem solving.

\section{Benchmark Suite}

\subsection{Core Competencies Taxonomy of Scientific Intelligence}

To comprehensively evaluate scientific intelligence in large language models, we construct a modality-aware taxonomy of seven core dimensions, classified into multimodal and text-only categories based on their input format and cognitive demands.
Multimodal benchmarks emphasize the integration and alignment of visual and text contents to meet the reasoning demands of real-world scientific workflows, while text-only benchmarks probe symbolic, conceptual, and generative reasoning capabilities solely through language.

\small
\begin{longtable}{@{}lllll@{}}
\caption{\centering Taxonomy of Benchmarks for Scientific Intelligence capabilities.}\\
\toprule
\textbf{Scientific Capability} & \textbf{Subject} & \textbf{Benchmark} & \textbf{Modality} & \textbf{Source}\\
\midrule
\endfirsthead
\toprule
\textbf{Scientific Capability} & \textbf{Subject} & \textbf{Benchmark} & \textbf{Modality} & \textbf{Venue}\\
\midrule
\endhead
\midrule
\endfoot
\bottomrule
\endlastfoot

Scientific Multimodal
Perception & Life Science & \href{https://huggingface.co/datasets/BoKelvin/SLAKE}{SLAKE} & Image & ISBI 21\\
\midrule

Scientific Multimodal Reasoning & Earth Science & \href{https://huggingface.co/MSEarth}{MSEarth} & Image & arXiv 25\\
\midrule

\multirow{4}{*}{Scientific Multimodal Understanding} & Multidisciplinary & \href{https://huggingface.co/datasets/PrismaX/SFE}{SFE} & Image & NeurIPS 25 \\
& Earth Science & \href{https://huggingface.co/datasets/initiacms/OmniEarth-Bench}{OmniEarth} & Image & arXiv 25\\
& Life Science & \href{https://huggingface.co/datasets/foreverbeliever/OmniMedVQA}{OmniMedVQA} & Image & CVPR 24\\
& Physics & \href{https://huggingface.co/datasets/Cloudriver/PhyX}{PhyX} & Image & arXiv 25\\
\midrule

\multirow{12}{*}{Scientific Knowledge Understanding}
& Chemistry & \href{https://huggingface.co/datasets/jablonkagroup/ChemBench}{ChemBench} & Text & Nat Chem 25\\
& Chemistry & \href{https://huggingface.co/datasets/AI4Chem/ChemBench4K}{ChemBench4K} & Text & arXiv 25\\
& Chemistry & \href{https://huggingface.co/datasets/osunlp/SMolInstruct}{LLM4Chem} & Image & COLM 24\\
& Earth Science & \href{https://huggingface.co/datasets/Rose-STL-Lab/ClimaQA}{ClimaQA} & Text & ICLR 25\\
& Earth Science & \href{https://huggingface.co/datasets/PrismaX/Earth-Silver}{EarthSE} & Text & arXiv 25\\
& Life Science & \href{https://huggingface.co/datasets/tsynbio/ProteinLMBench}{ProteinLMBench} & Text & BIBM 24\\
& Life Science & \href{https://huggingface.co/datasets/GreatCaptainNemo/BioProBench}{BioProbench} & Text & arXiv 25\\
& Materials Science & \href{https://huggingface.co/datasets/heegyu/mascqa}{MaScQA} & Text & Digit Discov 24\\
& Life Science & \href{https://huggingface.co/datasets/GENTEL-Lab/TRQA}{TRQA} & Text & arXiv 25\\
& Life Science & \href{https://github.com/hhnqqq/Biology-Instructions}{Biology-Instructions} & Text & ACL 25 \\
& Life Science & \href{https://huggingface.co/datasets/zjunlp/Mol-Instructions}{Mol-Instructions} & Text & ICLR 24 \\
& Life Science & \href{https://github.com/DeepGraphLearning/PEER_Benchmark}{PEER} & Text & NeurIPS 22 \\
\midrule

\multirow{2}{*}{Scientific Code Generation}
& Multidisciplinary & \href{https://huggingface.co/datasets/SciCode1/SciCode}{SciCode} & Text & NeurIPS 24 \\
& Astronomy & \href{https://huggingface.co/datasets/sebajoe/AstroVisBench}{AstroVisBench} & Image & NeurIPS 25 \\
\midrule

\multirow{2}{*}{Scientific Symbolic Reasoning}
& Physics & \href{https://huggingface.co/datasets/weidawang/CMPhysBench}{CMPhysBench} & Text & arXiv 25\\
& Physics & \href{https://huggingface.co/datasets/desimfj/PHYSICS}{PHYSICS} & Text & arXiv 25\\
\midrule

\multirow{1}{*}{Science Hypothesis Generation}
& Multidisciplinary & \href{https://huggingface.co/datasets/ankilok/Researchbench}{ResearchBench} & Text & ICML 25
\label{tab:benchmark_taxonomy}
\end{longtable}

\textbf{Scientific Multimodal Perception} captures a model's ability to detect and localize scientifically meaningful entities from multimodal input.Unlike general visual perception, scientific perception requires identifying scientific structures such as organs in CT or MRI scans or chemically relevant patterns the images.
\textbf{Scientific Multimodal Understanding} represents the multimodal capability to extract and interpret structured scientific information from visual elements where the images themselves carry a high degree of scientific specificity. Benchmarks like SFE fall into this category where they require the model to align scientific symbols, notations and visual encodings with domain knowledge, making this capability essential for scientific workflows.
\textbf{Scientific Multimodal Reasoning} refers to the model's ability to integrate visual and textual modalities to support coherent scientific inference. This capability goes beyond recognizing modality-specific patterns. It emphasizes cross-modal grounding, multi-step inference and domain-aware reasoning strategies. A key facet involves Chain-of-Thought (CoT) style reasoning, where models articulate intermediate steps when answering complex science questions.
Within text-only benchmarks, we categorize benchmarks according to four key scientific capabilities. The first is \textbf{Scientific Knowledge Understanding}, which assesses a model's grasp of domain-specific concepts and factual relationships across disciplines such as Chemistry, Earth Science and Life Science. The second concerns \textbf{Scientific Code Generation} involving algorithmic comprehension and code generation tasks that demand precise mapping from scientific descriptions to executable logic. The third focuses on \textbf{Scientific Symbolic Reasoning} which targets a model's ability to manipulate equations, units, and structured scientific notations. This ability is particularly critical in disciplines such as physics, where symbolic representation plays a central role in modeling physical systems and deriving formal solutions. Finally, we include \textbf{Science Hypothesis Generation} where models engage in abductive inference and explanatory synthesis under open-ended or minimally structured prompts.

\subsection{Scientific Discipline Coverage}
A key design principle of our benchmark suite is to ensure comprehensive coverage of the major scientific disciplines where large language models are expected to demonstrate expert-level reasoning. Rather than evaluating models through isolated subject-specific tasks, our framework spans the full landscape of natural sciences, including \textbf{Life Science, Chemistry, Earth Science, Physics, Astronomy and Materials Science}.

Each discipline is represented through benchmarks that reflect not only factual or textbook-level knowledge, but also procedural reasoning, mechanistic interpretation, and context-dependent application. For example, ProteinLMBench~\cite{shen2024fine} and TRQA-lit~\cite{zhang2025origene} capture biomolecular and biomedical reasoning, requiring the integration of protein sequence understanding with biological function and therapeutic context. ChemBench~\cite{Mirza2025} and MaScQA~\cite{zaki2024mascqa} evaluate higher-order chemical and materials reasoning, including thermodynamics, phase transitions, and structure–property analysis. ClimaQA~\cite{manivannan2024climaqa} and EarthSE~\cite{xu2025earthse} focus on earth system, climate science, and geospatial interpretation, while PHYSICS~\cite{zheng2025scaling} and CMPhysBench~\cite{wang2025cmphysbench} emphasize fundamental physical laws, mathematical modeling, and symbolic derivations.

This broad disciplinary coverage enables the evaluation of not only knowledge-centric tasks, but also the cognitive heterogeneity across scientific domains where reasoning formats, cognition levels, modality dependencies, and knowledge structures differ substantially. As such, our benchmark suite offers a more faithful and holistic view of how LLMs generalize across scientific fields, rather than excelling in narrow domains.

\subsection{Expert-Aligned Benchmark Construction}
\subsubsection{Principles of Expert‑Aligned Benchmark Design}
To ensure the benchmark suite faithfully evaluates the scientific capabilities of LLMs, we adopt a construction paradigm grounded in domain expertise, cognitive coverage and procedural transparency. The benchmarks aims to represent core scientific workflows through rigorously selected tasks that reflect pressing and high‑impact scientific questions of contemporary importance.

We conduct a multi-round consultation with domain experts from diverse scientific fields including chemistry, earth science, life science, materials science, and physics. These experts are invited to propose benchmark tasks reflective of authentic research challenges they encounter ranging from climate assessment, protein function inference, thermodynamic reasoning in material design and scientific code generation. Proposals are evaluated based on their fidelity to real-world scientific reasoning and alignment with high-priority questions in specific scientific domains.

Following the open-ended proposal phase, a second-stage selection process is implemented to identify core benchmarks that satisfied below criteria:
\begin{enumerate}
    \item \textbf{Scientific Validity}: Tasks must be grounded in real scientific content and reasoning, avoiding mechanical factual memory. For example, questions like chemical reaction pathways or protein structure interpretation are preferred over superficial concept definitions.
    \item \textbf{Expert Calibration}: Each benchmark undergoes manual verification and calibration by domain experts, who validate the correctness of task formulations, solution rationales, and scoring criteria.
    \item \textbf{Capability Coverage}: Selected benchmarks must collectively span the five core dimensions of scientific intelligence—Scientific Knowledge Understanding, Scientific Code Generation, Symbolic Reasoning, Hypothesis Generation, and Diagram Understanding—ensuring that the evaluation reflects both analytical depth and reasoning breadth.
    \item \textbf{Modality and Task Diversity}: The suite covers multiple modalities (text, diagrams, molecular structures, protein sequences, scientific plots, radiological imagery, etc.) and multiple task formats (multiple-choice, free-form generation, code execution, document analysis), thereby capturing the multimodal and procedural nature of real scientific workflows.
    \item \textbf{Community Recognition}: Benchmarks are preferably endorsed by the broader scientific or industrial community at domain-leading conferences, or released by reputable research groups. This ensures the benchmarks' credibility, relevance and alignment with community-validated standards.
\end{enumerate}

\subsubsection{Benchmark Overview}
To assess the scientific intelligence across diverse modalities and tasks, we curated a suite of expert-aligned benchmarks that reflect real-world scientific workflows. Each benchmark is designed to evaluate specific dimension of scientific intelligence. 
Additionally, these benchmarks span a broad spectrum of scientific disciplines including chemistry, earth sciences, life sciences, materials science, and astrophysics.

We provide a systematic evaluation of scientific intelligence from two dimensions: text-only and multimodal evaluation. The text-only benchmarks focus on evaluating a model’s ability to comprehend, reason over, and generate scientific content using purely textual input. 
In contrast, the multimodal benchmarks introduce scientific problems where text and visual inputs are both required. Such settings simulate real-world scientific scenarios where visual and textual reasoning must be integrated, enabling a more complete evaluation of the model’s ability to engage with complex scientific artifacts. We provide detailed descriptions of each benchmark including their disciplinary coverage, task design, and alignment with scientific capabilities below.

Our first release evaluates models on a curated subset of {\ProjectName} benchmark pool. Detailed dataset description are referred to Appendix~\ref{app:bench_details}. At a glance, the present suite comprises:
\begin{itemize}
    \item \emph{Text-only evaluation:} centric benchmarks (e.g.\ ChemBench, MaScQA, and
    ProteinLMBench), two specialised reasoning sets for code generation (SciCode, AstroVisBench)
    and symbolic manipulation (CMPhysBench, PHYSICS).  
    \item \emph{Multimodal evaluation:} vision–language benchmarks whichi are MSEarth, SLAKE and
    SFE, requiring joint reasoning over scientific figures and textual context.
\end{itemize}
Together these tasks span chemistry, earth science, life science, materials science, physics, 
and astronomy. The suite was selected through our expert-aligned construction workflow to ensure (i) high scientific validity, (ii) coverage of all seven capability dimensions, and (iii) diversity of modalities and question types.  

Table~\ref{tab:benchmark_taxonomy} summarizes the core benchmarks included in this release of {\ProjectName}. These benchmarks are selected based on their coverage of key scientific intelligence capabilities across diverse disciplines and modalities. Specifically, this table highlights the most representative and capability-aligned benchmarks that form the evaluation backbone for our first leaderboard release.
And we will continuously expand the coverage in future releases. This includes incorporating additional modalities, disciplines, and newly proposed agent tasks from the community.

\section{Evaluation Framework}
\begin{figure*}[t]
  \centering
  \includegraphics[width=\linewidth]{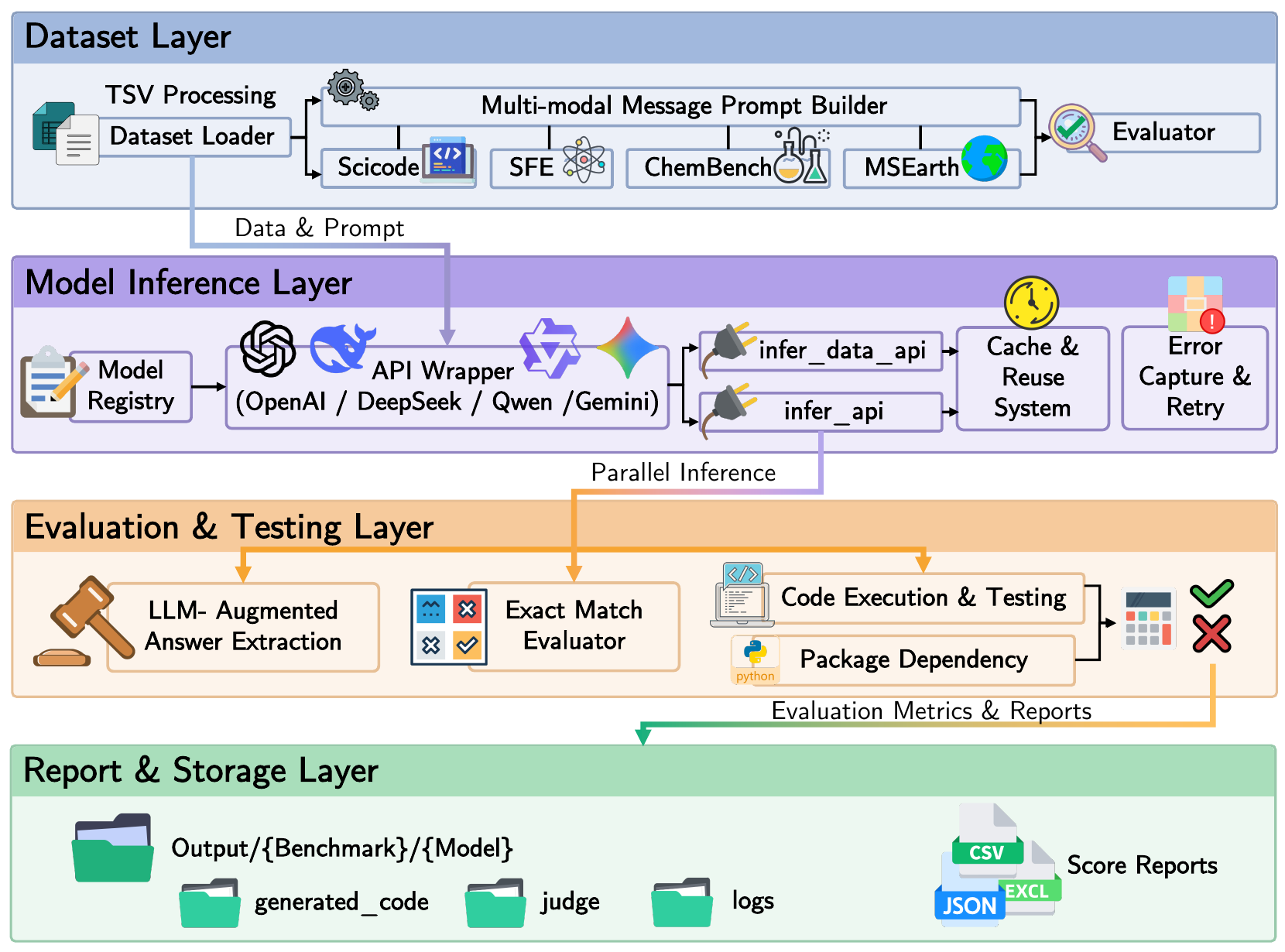}
  \caption{\centering Evaluation pipeline used in {\ProjectName}.}
  \label{fig:framework}
\end{figure*}

\subsection{Abstraction Layer}
The framework of {\ProjectName} is organised as four cooperating layers which are Dataset, Model Inference, Evaluation \& Testing, and Report \& Storage that together deliver an end-to-end, reproducible pipeline for multimodal scientific benchmarking.  Each layer's scope is deliberately kept concise with clearly defined boundaries so that researchers can extend one part of the stack without impacting the others.

\noindent \textbf{Dataset Layer.}
The Dataset Layer serves as the entry point for data ingestion and task specification. Dataset construction is handled through the \texttt{build\_dataset} routine, which maps a dataset identifier to its respective dataset class using centralized registries (e.g., \texttt{supported\_video\_datasets}, \texttt{supported\_text\_datasets}). Each dataset class inherits from either \texttt{TextBaseDataset}, \texttt{ImageBaseDataset}, or \texttt{VideoBaseDataset}, which provide unified interfaces for TSV or metadata loading, index normalization, and modality-specific data caching. Each dataset implements a custom \texttt{build\_prompt()} method that encapsulates raw task data such as questions, images, video frame paths, code snippets, and answer options into interleaved multi-modal messages. This representation constitutes the atomic unit for model-level inference.

\noindent \textbf{Model Inference Layer.}
The Model Inference Layer mediates between structured prompts and model outputs. Model instantiation is handled by \texttt{build\_model\_from\_config}, which resolves model metadata from \texttt{supported\_VLM}. Each model object exposes a unified \texttt{.generate(message, dataset)} interface, abstracting the distinction between local inference (via vLLM or torch.distributed) and API-based cloud models (OpenAI, DeepSeek, Gemini, Anthropic). Inference workflows are orchestrated via \texttt{infer\_data()},  \texttt{infer\_data\_api()} and \texttt{infer\_data\_job\_video()}, which provide transparent support for batching, parallel token generation, retry or error tolerance, and partial progress restoration through \texttt{\_supp.pkl}.

\noindent \textbf{Evaluation \& Testing Layer.}
Once predictions are generated, dataset-specific evaluate methods perform capability-aligned scoring through a combination of exact matching, semantic retrieval, numerical scoring, code execution, and LLM-based judging function \texttt{evaluate()}. Matching utilities, multiple-choice answer extraction and general-purpose judging \texttt{build\_judge()} provide deterministic and LLM-augmented evaluation paths. Code-execution tasks (e.g., Scicode) invoke sandboxed Python environments to verify visual output fidelity and computational correctness.

\noindent \textbf{Report \& Storage Layer.}
The Report \& Storage Layer ensures reproducibility and transparent logging. All predictions, logs, reasoning traces, metadata, and evaluation results follow a structured file convention. Helper functions (e.g., \texttt{get\_pred\_file\_path()}, \texttt{prepare\_reuse\_files()}, \texttt{get\_intermediate\_file\_path()}) ensure consistency across model runs. Final evaluation metrics are serialized in CSV, JSON, or XLSX formats based on benchmark requirements, facilitating both longitudinal comparison and leaderboard hosting.

\subsection{Unified interface for prompt construction and prediction}
{\ProjectName} is developed on the top of the VLMEvalKit~\cite{duan2024vlmevalkit} code base, preserving its modular abstractions while introducing extensions for scientific multimodal inputs, discipline-aware prompt construction and capability-oriented evaluation
A central design goal of {\ProjectName} is to provide a unified interface through which datasets, models, and evaluators interact in a modality-agnostic manner. To achieve this, the framework establishes a standardized prompt construction and prediction interface that applies consistently to text-only tasks, image-based visual reasoning, and multi-modal scientific problems involving diagrams, code snippets, symbolic expressions, molecular structures, and geospatial imagery. This interface is implemented at both the dataset and model abstraction layers, enabling fully unified end-to-end execution without requiring per-dataset or per-model procedural handling.

On the dataset side, every dataset class inherits from a base abstraction such as TextBaseDataset, ImageBaseDataset, or VideoBaseDataset, each of which exposes a common set of required functions. Among them, the \texttt{.build\_prompt() }method is the core entrypoint responsible for converting a structured sample from the dataset (e.g., TSV row containing question text, answer options, image encoding, or code references) into a standardized multi-modal message representation. A multi-modal message is represented as an ordered list of typed content segments, where each segment explicitly declares both its modality and payload. 
\begin{lstlisting}[language=Python]
dict(type='text', value=text)
dict(type='image', value=tgt_path)
\end{lstlisting}
This explicit specification ensures that models can interpret consistently. If a dataset supports custom instruction formats, such as SFE’s discipline-aware prompt templates for multiple-choice, exact-match, or open-ended questions, these are applied within \texttt{.build\_prompt() } while still adhering to the unified output message schema. The interface also accommodates advanced message packing for video or sequential images when activated in \texttt{infer\_data\_api()}.

Additionally, datasets provide a \texttt{.evaluate()} method as a standardized evaluator interface, which get model predictions from the storage layer and applies deterministic or judge-assisted scoring. Depending on the dataset type, the evaluation pipeline may involve exact matching, exact matching, choice extraction, code execution for scientific programming tasks, or LLM-based scoring via \texttt{build\_judge()}. Despite these internal variations, the signature of the \texttt{evaluate()} function remains constant across datasets, ensuring compatible execution. Supporting utility functions such as \texttt{display()} and \texttt{dump\_image()} further contribute to this unified interface by enabling consistent visual inspection, debugging, and handling of base64-encoded or remote images, ensuring that dataset parsing and integrity checks are conducted through shared mechanisms.

On the model side, the framework enforces a single inference interface through \texttt{.generate()}, regardless of whether the model is accessed via API endpoints (e.g., GPT-4o, Gemini, DeepSeek) or runs locally through vLLM or PyTorch-based implementations. All inference functions including \texttt{infer\_data()},  \texttt{infer\_data\_api()} and \texttt{infer\_data\_job\_video()} construct messages from dataset prompts and invoke \texttt{model.generate(message=..., dataset=..., **kwargs)} in a unified manner. This uniform invocation mechanism abstracts away differences in backend execution, request formatting, batching, retry handling, or temperature sampling. Moreover, models may optionally override prompt formatting by checking \texttt{use\_custom\_prompt(dataset\_name)}, while still maintaining conformity to the unified interface contract.

This unified interface design enables powerful decoupling between dataset logic, model execution, and evaluation strategy. Researchers can incorporate new datasets by providing only \texttt{.build\_prompt()} and \texttt{.evaluate()}, without modifying inference or scoring pipelines. Models can be swapped freely by exposing a \texttt{.generate()} method that accepts standardized multi-modal messages. Inference scripts can construct prompts, dispatch model generation, transfer results, and trigger evaluation without conditional branching or dataset-specific logic. As a result, the entire pipeline is fully modular, reproducible and extensible, highlighting the architectural robustness of {\ProjectName}.

\subsection{Capability-Oriented Evaluation}
To move beyond aggregate accuracy and better reflect the multifaceted nature of scientific intelligence, we adopt a capability-oriented evaluation paradigm. Rather than treating all benchmarks uniformly, we classify them into five core competency dimensions based on the underlying cognitive demands, task structure, and the knowledge representations required for successful problem solving. This allows each model to be evaluated not only on performance, but on what kind of scientific reasoning it is capable of. 
For each capability dimension, a model’s score is computed as the average performance across all benchmarks belonging to that capability, ensuring that evaluation is both comprehensive and domain-balanced.

\textbf{Scientific Multimodal Perception.} This dimension measures a model’s ability to detect and localise scientifically meaningful entities in visual inputs. Unlike generic perception, it targets domain-specific structures, e.g., organs in CT/MRI.

\textbf{Scientific Multimodal Understanding.} This dimension evaluates how well a model interprets visually encoded scientific information and aligns it with text. Tasks range from satellite environmental maps and annotated medical scans to molecular diagrams, demanding entity localisation, structural decoding, and fine-grained diagram-text alignment. 

\textbf{Scientific Multimodal Reasoning.} This dimension assesses a model’s ability to integrate visual and textual evidence to perform multi-step, domain-aware reasoning. Chain-of-Thought (CoT) articulation, cross-modal grounding, and disciplined scientific inference are required to reach correct answers.

\textbf{Scientific Code Generation.} This dimension evaluates a model’s capacity to translate scientific intent into executable computational procedures. \textit{AstroVisBench} and \textit{SciCode} are assigned to this category because their tasks require not only generating syntactically correct code, but also reasoning over domain-specific computational logic (e.g., astrophysics data processing, numerical simulation and scientific plotting). The score for this capability is computed as the average over these benchmarks, reflecting both semantic correctness and the ability to produce executable code that aligns with real-world scientific workflows and engineering practices.

\textbf{Scientific Symbolic Reasoning.} This dimension focuses on symbolic knowledge representation, formula manipulation, unit reasoning, and quantitative derivation. \textit{CMPhysBench} and \textit{PHYSICS} both evaluate these capacities using tasks that require algebraic transformation, dimensional consistency, and symbolic inference rather than mere textual recall. The capability score is derived from the average results across these symbolic reasoning benchmarks. These tasks are particularly representative of the unique reasoning challenges found in physics where symbolic representations are fundamental to problem solving.

\textbf{Science Hypothesis Generation.} \textit{ResearchBench} is the only benchmark dedicated to this capability and uniquely represents open-ended hypothesis formulation, scientific discourse planning, and research proposal synthesis. Unlike conventional QA tasks, it evaluates abductive reasoning, conceptual integration, novelty generation, and literature-grounded justification.

\textbf{Scientific Knowledge Understanding.} This dimension evaluates a model’s capacity to comprehend and reason over domain-specific scientific knowledge across a broad spectrum of disciplines. Benchmarks such as ProteinLMBench, MaScQA, ClimaQA, TRQA-lit, Earth-Silver, and ChemBench focus on concept interpretation, factual consistency, mechanistic understanding, property relations, and application-oriented reasoning grounded in real scientific contexts. These tasks typically involve integrating procedural knowledge, conceptual hierarchy, and disciplinary logic rather than symbolic derivation or general linguistic inference.
The score for this capability dimension is computed as the average performance across all benchmarks categorized under Scientific Knowledge Understanding.

\subsection{Evaluation Modes}
To ensure that model performance is assessed in a manner faithful to scientific rigor, our evaluation framework adopts a hybrid scoring paradigm that integrates \textbf{deterministic rule-based matching, semantic LLM-based judging, and execution-based verification}. This design accommodates the diverse answer formats and reasoning modalities present across scientific tasks ranging from symbolic problem solving and diagram interpretation to code synthesis and explanatory reasoning.

\textbf{Evaluation Methods.}
We employ two primary evaluation pipelines:
\begin{itemize}
    \item Natural-language matching: This is used for most knowledge, reasoning, and visualization oriented tasks. Model predictions are extracted and normalized via rule-based processing (e.g., option inference, unit normalization, numerical span extraction, or text canonicalization), and then compared against ground truth using dataset-specific metrics. For different task types, we adopt appropriate scoring schemes such as accuracy (MCQ) or relaxed numeric matching (scientific problem solving).
    \item Code–execution–based evaluation: For benchmarks such as SciCode that explicitly evaluate scientific code generation and algorithmic reasoning, the model’s output is interpreted as executable Python programs. The predicted code is stitched into functional scripts, dependencies are resolved, and official unit tests are executed to verify correctness. Scores are computed based on the number of passed test cases, reflecting not only syntactic validity but also functional correctness and engineering reliability.
\end{itemize}

\textbf{Question Format Handling.}
To accommodate heterogeneous formats across benchmarks, we adopt differentiated evaluation strategies:
\begin{itemize}
    \item Multiple-choice questions (MCQ): Predictions are mapped to option labels through rule-based extraction. If the model outputs free-form text, a secondary semantic alignment step infers the intended option (A/B/C/D). Accuracy is reported, optionally with per-category results.
    \item Fill-in-the-blank / cloze questions: Extracted responses are reduced to core information (numbers, chemical formulas, single phrases) using handcrafted parsing rules. Answers are scored based on exact match or relaxed matching with semantic normalization.
    \item Open-ended / free-form questions: For explanation questions, descriptive captioning, or scientific hypothesis formulation, we compute string- and semantics-based metrics (e.g., BLEU, VQA score, or semantic overlap). In certain benchmarks, optional qualitative grading (e.g., correctness, reasoning soundness) is also permitted using LLM-as-judge.
\end{itemize}

\textbf{Judgment Strategies.}
To accurately evaluate answers requiring semantic equivalence, contextual reasoning, or scientific justification, we adopt two scoring strategies:
\begin{itemize}
    \item Rule-based strategy: Deterministic string normalization, regular expression extraction, unit conversion, and option inference handle most cases involving structured answers or symbolic formats. 
    \item LLM Evaluation Strategy: When rule-based approaches fail—particularly for free-form responses, complex reasoning explanations, or ambiguous mappings, we optionally invoke verification models (primarily closed-source SOTA models and \ProjectName predominantly using GPT series) to assess semantic equivalence or correctness.
\end{itemize}

\section{Evaluation Results}
\begin{figure*}[h]
  \centering
  \includegraphics[width=0.8\linewidth]{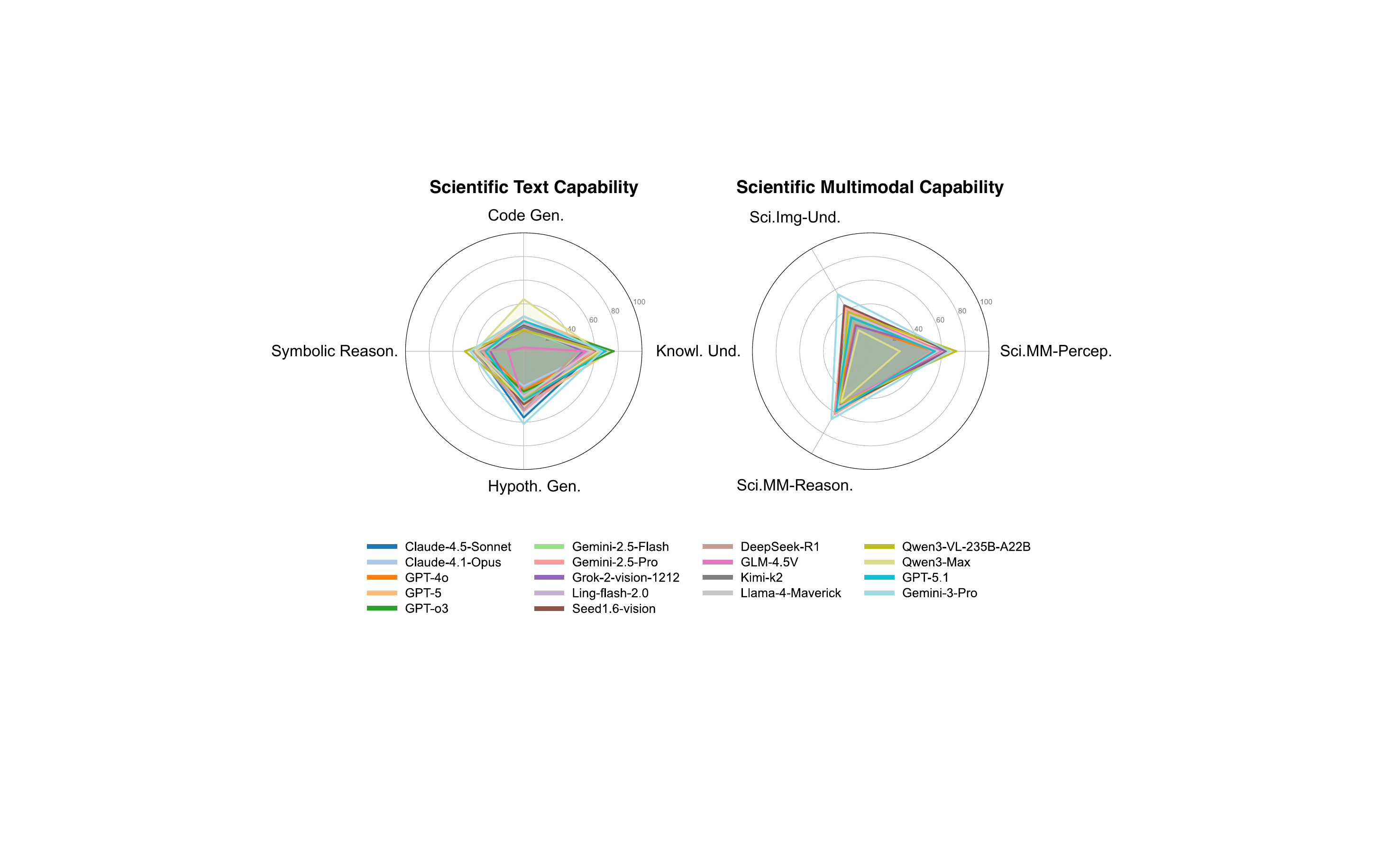}
  \caption{Large-language-model (LLM) scientific capabilities (left) versus multimodal-language-model (MLLM) scientific capabilities (right) comparing the evaluated models on the SciEval leaderboard. Each axis reports the score (0 – 100) for one capability or scientific field; concentric rings mark 20 intervals up to the outer 100 score. }
  \label{fig:results}
\end{figure*}

We present a comprehensive evaluation of contemporary large language models (LLMs) and multimodal large language models (MLLMs) using the {\ProjectName} evaluation suite. Fig.~\ref{fig:results} summarizes model performance across the core dimensions of scientific text capability (left) and scientific multimodal capability (right), with each axis reporting normalized scores in the range of 0–100. Rather than collapsing results into a single aggregate metric, we analyze model behavior along individual cognitive and modality-specific dimensions, enabling a more fine-grained characterization of scientific intelligence.

Fig.~\ref{fig:results} reveals several systematic patterns that align with the quantitative leaderboards while providing additional insights.  First, models achieve their highest scores on Scientific Knowledge Understanding across the board of scientific text ability, while Code Generation and Symbolic Reasoning remain substantially weaker. This imbalance underscores that contemporary models have largely solved factual recall but still struggle with executable logic and equation-level manipulation, two skills that remain essential for modern computational science. 
Second, proprietary models which are Gemini 3 Pro and GPT-5 form the widest and most balanced polygons, reflecting the highest average scores and the most even performance.  Notably, the open-source Qwen3-VL-235B-A22B achieves scores that approach those of the leading proprietary models, indicating that some open-source systems now perform competitively across multiple scientific competencies.

Most evaluated MLLMs demonstrate comparatively strong performance in scientific multimodal perception, reflecting progress in aligning visual features with textual semantics, particularly for entity recognition and diagram-level grounding. This suggests that current vision–language alignment is sufficient for basic, surface-level scientific perception, but remains limited in depth and robustness.

A closer comparison between GPT-5 and its follow-up GPT-5.1 indicates that scores decrease across nearly every axis, indicating that the latest iteration prioritizes incremental alignment refinements over new scientific competence. 
Finally, the highest scores are still achieved by general-purpose models such as Gemini 3 Pro and GPT-o3 rather than by any domain-specific, science-tuned models.

\renewcommand{\arraystretch}{1.25}
\small
\setlength{\tabcolsep}{4pt}

\begin{longtable}{@{}p{4.2cm}cccccc@{}}
\caption{Evaluation of large language models across four scientific text capability benchmarks: Scientific Knowledge Understanding (Knowl. Und.), 
Scientific Code Generation (Code Gen.), 
Scientific Symbolic Reasoning (Symbolic Reason.), 
and Scientific Hypothesis Generation (Hypoth. Gen.). }
\label{tab:text_ability} \\
\toprule
\textbf{Model} & \textbf{Knowl.\ Und.} & \textbf{Code Gen.} & \textbf{Symb. Reason.} & \textbf{Hypoth.\ Gen.} &  \textbf{Overall} \\
\midrule
\endfirsthead

\toprule
\textbf{Model} & \textbf{Knowl.\ Und.} & \textbf{Code Gen.} & \textbf{Symb. Reason.} & \textbf{Hypoth.\ Gen.} & \textbf{Overall} \\
\midrule
\endhead

\midrule
\endfoot

\bottomrule
\endlastfoot

\rowcolor{CloseBG}\multicolumn{6}{@{}c}{\textbf{Closed-Weight LLMs}}\\
\midrule
Claude 4.5 Sonnet~\cite{anthropic2025claude45}  & 60.67 & 21.73 & 40.36 & \silver{56.10} & 44.72 \\
Claude 4.1 Opus~\cite{anthropic2025claude41}     & 60.87 & 25.32 & 38.69 & 29.47 & 38.58 \\
GPT-5.1~\cite{openai2025gpt51}            & \copper{69.23} & 25.63 & 32.44 & 41.45 & 42.19 \\
GPT-5~\cite{openai2025gpt5}       & \silver{74.05} & \copper{29.21} & 39.91 & 45.67 & \copper{47.21} \\
GPT-4o~\cite{openai2024gpt4o}      & 60.84 & 17.67 & 32.09 & 33.04 & 35.91 \\
GPT-o3~\cite{openai2025o3}             & \gold{76.05} & 25.26 & 38.14 & 34.14 & 43.40 \\
Gemini-3-Pro~\cite{google2023gemini3blog}       & 66.06 & \silver{29.57} & \silver{45.19} & \gold{61.51} & \gold{50.58} \\
Gemini-2.5-Flash~\cite{google2025gemini25flash}   & 50.46 & 18.28 & 32.07 & 40.86 & 35.42 \\
Gemini-2.5-Pro~\cite{google2025gemini25pro}     & 59.34 & 24.77 & 34.96 & \copper{50.73} & 42.45 \\
Grok-2-vision-1212~\cite{xai2025grok2vision1212} & 50.14 & 20.60 & 28.21 & 49.63 & 37.14 \\
Seed1.6-vision~\cite{seed2025vision16}     & 65.78 & 21.49 & 39.24 & 45.00 & 42.88 \\
\midrule
\rowcolor{OpenBG}\multicolumn{6}{@{}c}{\textbf{Open-Weight LLMs}}\\
\midrule
GLM-4.5V~\cite{vteam2025glm45vglm41vthinkingversatilemultimodal}           & 52.78 & 3.24  & 13.43 & 42.23 & 27.92 \\
Ling-flash-2.0~\cite{li2025every}     & 53.39 & 25.60 & 37.98 & 50.29 & 41.81 \\
DeepSeek-R1~\cite{guo2025deepseek}        & 45.17 & 2.35  & \copper{42.80} & 49.73 & 35.01 \\
Kimi-k2~\cite{team2025kimi}            & 62.49 & 20.86 & 38.59 & 42.28 & 41.06 \\
Llama 4 Maverick~\cite{meta2025llama}   & 57.22 & 18.26 & 38.97 & 38.31 & 38.19 \\
Qwen3-VL-235B-A22B~\cite{bai2025qwen3vltechnicalreport} & 65.98 & 18.00 & \gold{49.93} & 40.62 & 43.63 \\
Qwen3-Max~\cite{yang2025qwen3}          & 63.14 & \gold{43.97} & {41.04} & 42.12 & \silver{47.57} \\
\end{longtable}

\begin{figure}[t]
  \centering
  \includegraphics[width=\linewidth]{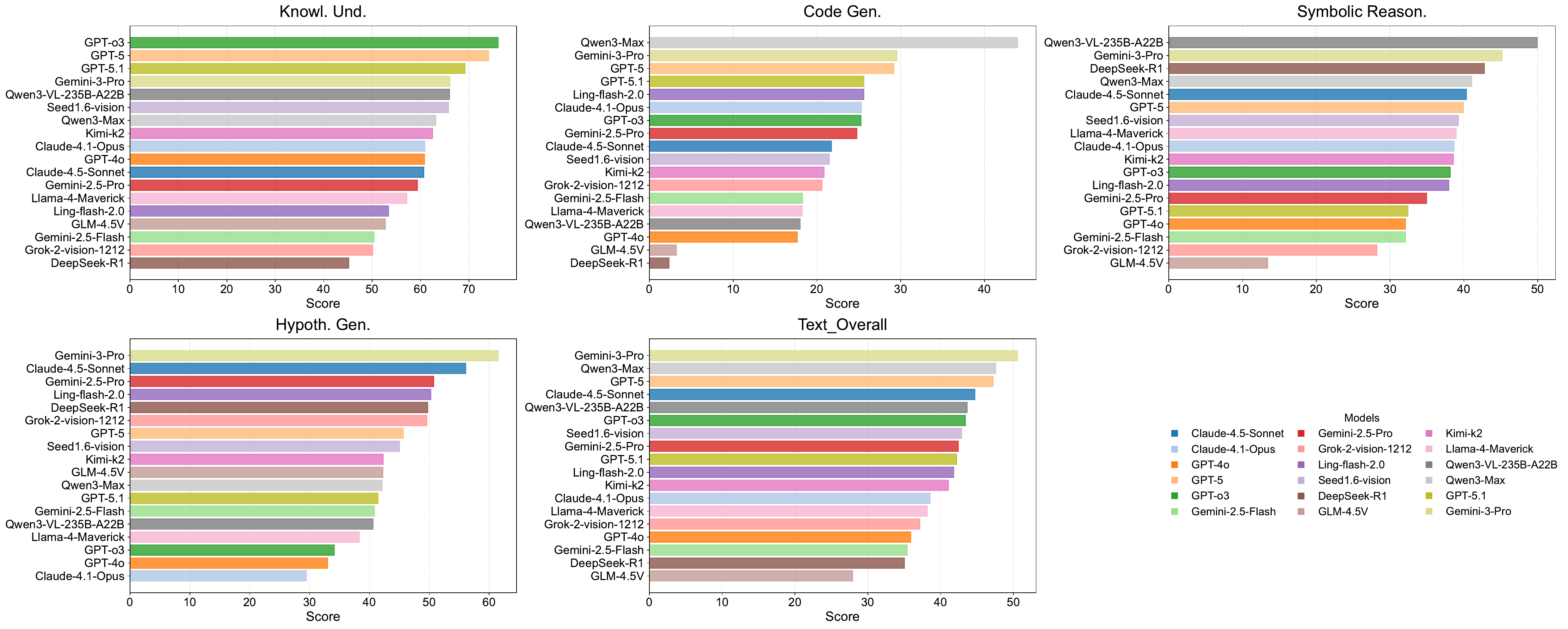}
  \caption{Model scores on four text-only capacities: Scientific Knowledge Understanding (Knowl. Und.), Scientific Code Generation (Code Gen.), Scientific Symbolic Reasoning (Symbolic Reason.), Scientific Hypothesis Generation (Hypoth. Gen.), and their mean (Text Overall). }
  \label{fig:Sci.text}
\end{figure}

\begin{figure}[t]
  \centering
  \includegraphics[width=0.9\linewidth]{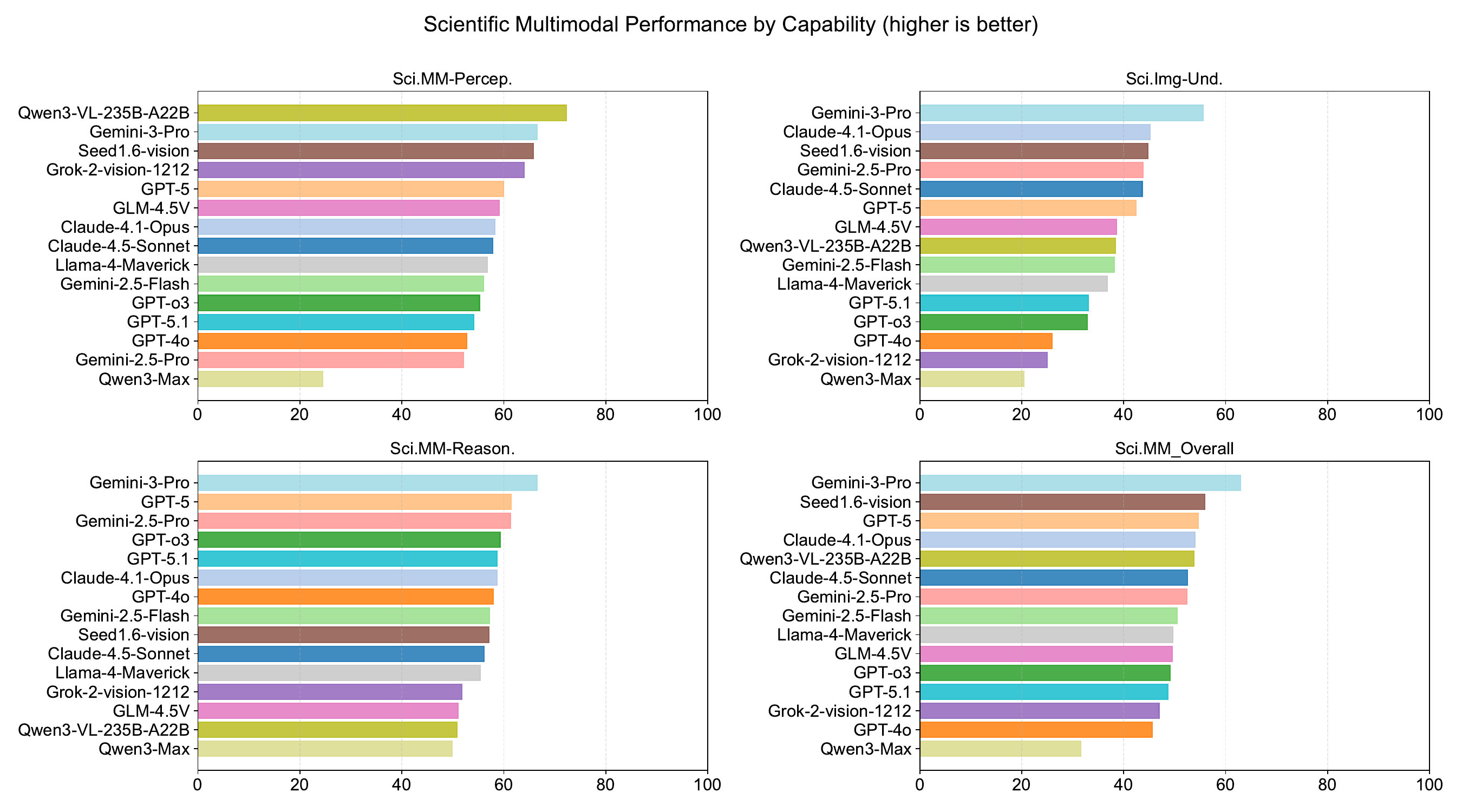}
  \caption{Model scores on three multimodal competencies: Scientific MM Perception (Sci. MM Perception), Scientific Multimodal Understanding (Sci.MM-Und.), and Scientific MM Reasoning (Sci.MM-Reason.) together with their average (Sci.MM-Overall). }
  \label{fig:Sci.MM}
\end{figure}

\paragraph{Scientific Text Capabilities}
We first assess the models based on four scientific text core capabilities: Scientific Knowledge Understanding (Knowl. Und.), Scientific Code Generation (Code Gen.), Scientific Symbolic Reasoning (Symbolic Reasoning), Scientific Hypothesis Generation (Hypoth. Gen.) as shown in Table~\ref{tab:text_ability}.

Models such as Gemini-3-Pro, GPT-5 and Qwen3-Max rank among the top performers overall, particularly in Scientific Knowledge Understanding and Hypothesis Generation. Their strong performance reflects broad scientific coverage and effective abstraction over domain concepts. 
Across the board, Gemini 3 Pro demonstrates the strongest overall capability, ranking first in four of five dimensions. It achieves the outstanding performance in Hypothesis Generation (61.51), Symbolic Reasoning (45.19), and Code Generation (29.57). 
GPT-5, GPT-o3, and Claude 4.5 Sonnet form a competitive group with balanced capability profiles. GPT-5 attains the second-highest scores in Knowledge Understanding (74.05), while GPT-o3 shows strongest performance in Knowledge Understanding (76.05). Claude models, although slightly behind Gemini-series in higher-order reasoning, demonstrate stable performance in Hypothesis Generation (56.10) and Symbolic Reasoning (40.36).
And these leading models exhibit clear limitations. While their knowledge understanding scores are high, performance in Code Generation and Symbolic Reasoning remains substantially lower, indicating that strong declarative knowledge does not directly translate into reliable formal or executable reasoning. In particular, symbolic manipulation and program correctness continue to pose challenges, even for the best-performing systems. Scientific Code Generation remains the weakest competency for all evaluated models, underscoring the gap between text-based reasoning and executable scientific problem-solving. Even top-performing models achieve relatively low scores (Qwen3-Max: 43.97, Gemini-3-Pro: 29.57, and GPT-5: 29.21), suggesting that current LLMs struggle with implementing algorithmic structures and translating scientific logic into runnable code.

Beyond absolute performance levels, different scientific text capabilities exhibit markedly different discriminative power across models. In particular, Code Generation shows the largest performance variance, sharply separating top-performing models from the rest of the leaderboard. While a small number of models demonstrate relatively strong executable reasoning ability, the majority exhibit consistently low scores, indicating that code-centric scientific reasoning remains a highly selective capability rather than a broadly acquired one.
It is noteworthy that Code Generation performance across different models often correlates positively with Symbolic Reasoning capabilities. Systems demonstrating superior Code Generation capabilities typically exhibit relatively higher Symbolic Reasoning scores, while models that perform poorly in symbolic operations often yield unsatisfactory results on executable tasks. This consistency suggests that both capabilities rely on shared foundational abilities, such as formal abstraction, step-by-step logical consistency, and tolerance for strict correctness constraints.

Fig.~\ref{fig:Sci.text} further corroborates these plain-text test results: the bar chart reveals that scientific knowledge comprehension exhibits tight clustering—with GPT-o3, GPT-5, and Gemini-3-Pro occupying the top three positions—while code generation and symbolic reasoning show the greatest dispersion. Qwen3-Max and Qwen3-VL-235B-A22B lead in these domains, but most other models experience a rapid decline in performance.

Remarkably, Qwen3-Max stands out among all models. Its performance is competitive with, and in some cases comparable to, the strongest proprietary systems across scientific text capabilities. This advantage is particularly shown in Code Generation, where Qwen3-Max consistently achieves the highest scores, clearly separating itself from other models.

\renewcommand{\arraystretch}{1.25}
\small
\setlength{\tabcolsep}{4pt}

\begin{longtable}{@{}p{4.5cm}cccc@{}}
\caption{Evaluation of LLMs on scientific multimodal capabilities, including 
Scientific Multimodal Perception (Sci.MM-Percep.), 
Scientific Multimodal Understanding (Sci.MM-Und.), 
and Scientific Multimodal Reasoning (Sci.MM-Reason.).}
\label{tab:multimodal_ability} \\
\toprule

\textbf{Model} & \textbf{Sci.MM-Percep.} & \textbf{Sci.MM-Und.} & \textbf{Sci.MM-Reason.} & \textbf{Overall} \\
\midrule
\endfirsthead

\toprule
\textbf{Model} & \textbf{Sci.MM-Percep.} & \textbf{Sci.MM-Und.} & \textbf{Sci.MM-Reason.} & \textbf{Overall} \\
\midrule
\endhead

\midrule
\endfoot

\bottomrule
\endlastfoot

\rowcolor{CloseBG}\multicolumn{5}{@{}c}{\textbf{Closed-Weight MLLMs}}\\
\midrule

\makecell[{{p{4cm}}}]{Claude 4.5 Sonnet}  & 57.87 & 43.64 & 56.11 & 52.54 \\
\makecell[{{p{4cm}}}]{Claude 4.1 Opus}     & 58.25 & \silver{45.19} & 58.66 & 54.03 \\
\makecell[{{p{4cm}}}]{GPT-5.1}            & 54.10 & 33.05 & 58.73 & 48.63 \\
\makecell[{{p{4cm}}}]{GPT-5}              & 59.94 & 42.44 & \silver{61.46} & \copper{54.61} \\
\makecell[{{p{4cm}}}]{GPT-4o}             & 52.78 & 25.93 & 57.97 & 45.56 \\
\makecell[{{p{4cm}}}]{GPT-o3}             & 55.23 & 32.84 & 59.27 & 49.11 \\
\makecell[{{p{4cm}}}]{Gemini-3-Pro}       & \silver{66.54} & \gold{55.62} & \gold{66.49} & \gold{62.88} \\
\makecell[{{p{4cm}}}]{Gemini-2.5-Flash}   & 55.98 & 38.20 & 57.22 & 50.47 \\
\makecell[{{p{4cm}}}]{Gemini-2.5-Pro}     & 52.12 & 43.76 & \copper{61.28} & 52.39 \\
\makecell[{{p{4cm}}}]{Grok-2-vision-1212} & 64.00 & 25.04 & 51.76 & 46.93 \\
\makecell[{{p{4cm}}}]{Seed1.6-vision}     & \copper{65.79} & \copper{44.75} & 57.11 & \silver{55.88} \\

\midrule
\rowcolor{OpenBG}\multicolumn{5}{@{}c}{\textbf{Open-Weight MLLMs}}\\
\midrule
\makecell[{{p{4cm}}}]{GLM-4.5V}           & 59.10 & 38.57 & 51.04 & 49.57 \\
\makecell[{{p{4cm}}}]{Llama 4 Maverick}   & 56.74 & 36.83 & 55.39 & 49.65 \\
\makecell[{{p{4cm}}}]{Qwen3-VL-235B-A22B} & \gold{72.29} & 38.35 & 50.83 & 53.82 \\
\makecell[{{p{4cm}}}]{Qwen3-Max}          & 24.51 & 20.40 & 49.86 & 31.59 \\
\end{longtable}

\paragraph{Scientific Multimodal Capabilities}
Table~\ref{tab:multimodal_ability} reports performance on scientific multimodal capabilities, evaluating models across Scientific Multimodal Perception, Scientific Multimodal Understanding, and Scientific Multimodal Reasoning.
Clear stratification emerges among models. Gemini-3-Pro consistently achieves the strongest overall multimodal performance, maintaining relatively balanced scores across perception, understanding, and reasoning. GPT-5 and Seed1.6-Vision follow closely, exhibiting strong perception and moderate reasoning ability but still showing noticeable degradation at the understanding level. 

A particularly instructive contrast is observed within the Qwen family. Qwen3-VL-235B-A22B achieves the highest scores in multimodal perception, yet this advantage does not translate into corresponding gains in understanding or reasoning, where performance drops sharply. Conversely, Qwen3-Max, which excels in scientific text capabilities, ranks near the bottom in multimodal settings. This divergence highlights a key limitation of current systems: strong visual grounding or strong language reasoning alone is insufficient for scientific multimodal intelligence without deeper semantic integration across modalities. Overall, multimodal reasoning exhibits the greatest variance across models and thus provides the strongest discriminative signal, while perception-level capability is comparatively saturated and less indicative of advanced scientific competence.

Fig~\ref{fig:Sci.MM} summarizes multimodal performance along three axes. Scientific-MM Perception exhibits the largest variance, separating a small lead group which are Qwen3-VL-235B-A22B, Gemini-3-Pro, Seed 1.6-vision from the long tail. Scores drop appreciably on Image Understanding, confirming that reliable visual grounding does not guarantee correct extraction of scientific semantics. MM Reasoning is the most compact dimension with the exception of Gemini-3-Pro, which retains a modest margin, suggesting a shared ceiling once multi-step, figure-grounded inference is required. Among open-weight models, Qwen3-VL demonstrates that superior perception alone is insufficient without commensurate gains in higher-level multimodal cognition.

Taken together, Fig.~\ref{fig:Sci.text}-\ref{fig:Sci.MM} visually corroborate the benchmarking data: knowledge-based capabilities have largely become saturated, while formal reasoning abilities whether symbolic reasoning, executable reasoning, or multimodal reasoning, remain the core metric distinguishing top-tier models.

\section{Conclusion and Discussion}
This work introduces {\ProjectName}, an open-source toolkit and leaderboard for measuring scientific intelligence in large language and multimodal language models. By unifying high-quality benchmarks, seven capability dimensions, and six scientific disciplines within a single evaluation pipeline, {\ProjectName} exposes the scientific capability of LLMs.

The results suggest that future gains are less likely to come from scale alone and more from (i) execution-aware codes, (ii) symbolic and program-of-thought reasoning, and (iii) tighter integration of visual grounding with scientific semantics.

Our future planned work includes an agent track with tool-use and verification loops, expanded multimodal tasks featuring raw spectra, molecular graphs, and volumetric data, and quarterly community releases to incorporate newly proposed tasks and models.



\begingroup
\sloppy
\printbibliography[heading=bibintoc]
\endgroup

\clearpage
\appendix

\section{Appendix}

\subsection{Authors}
\textbf{Leading Authors}

Yiheng Wang, Yixin Chen

\textbf{Project Contributors}

Shuo Li, Yifan Zhou, Bo Liu, Hengjian Gao, Jiakang Yuan, Jia Bu, Wanghan Xu, Yuhao Zhou, Xiangyu Zhao, Zhiwang Zhou, Fengxiang Wang, Haodong Duan, Songyang Zhang

\textbf{Community Contributors}

Jun Yao, Han Deng, Yizhou Wang, Jiabei Xiao, Jiaqi Liu, Encheng Su, Yujie Liu, Weida Wang, Junchi Yao, Shenghe Zheng, Haoran Sun, Runmin Ma, Xiangchao Yan, Bo Zhang, Dongzhan Zhou, Shufei Zhang, Peng Ye, Xiaosong Wang, Shixiang Tang

\textbf{Corresponding Authors}

Wenlong Zhang, Lei Bai

\subsection{Full Evaluation Results Across Core Benchmarks}
\renewcommand{\arraystretch}{1.25}
\small
\setlength{\tabcolsep}{20pt}

\begin{longtable}{@{}p{4cm}ccc@{}}
\caption{\centering Evaluation of LLMs on scientific multimodal benchmarks: SLAKE, SFE, and MSEarth.}
\label{tab:multimodal_eval} \\
\toprule
\textbf{Model} & \textbf{SLAKE} & \textbf{SFE} & \textbf{MSEarth} \\
\midrule
\endfirsthead

\toprule
\textbf{Model} & \textbf{SLAKE} & \textbf{SFE} & \textbf{MSEarth} \\
\midrule
\endhead

\midrule
\endfoot

\bottomrule
\endlastfoot

\rowcolor{CloseBG}\multicolumn{4}{@{}c}{\textbf{Closed-Weight MLLMs}}\\
\midrule

\makecell[{{p{4cm}}}]{Claude 4.5 Sonnet} & 57.87 & 43.64 & 56.11 \\

\makecell[{{p{4cm}}}]{Claude 4.1 Opus} & 58.25 & 45.19 & 58.66 \\

\makecell[{{p{4cm}}}]{GPT-5.1} & 54.10 & 33.05 & 58.73 \\

\makecell[{{p{4cm}}}]{GPT-5} & 59.94 & 42.44 & 61.46 \\

\makecell[{{p{4cm}}}]{GPT-4o} & 52.78 & 25.93 & 57.97 \\

\makecell[{{p{4cm}}}]{GPT-o3} & 55.23 & 32.84 & 59.27 \\

\makecell[{{p{4cm}}}]{Gemini-3-Pro} & 66.54 & 55.62 & 66.49 \\

\makecell[{{p{4cm}}}]{Gemini-2.5-Flash} & 55.98 & 38.20 & 57.22 \\

\makecell[{{p{4cm}}}]{Gemini-2.5-Pro} & 52.12 & 43.76 & 61.28 \\

\makecell[{{p{4cm}}}]{Grok-2-vision-1212} & 64.00 & 25.04 & 51.76 \\

\makecell[{{p{4cm}}}]{Seed1.6-vision} & 65.79 & 44.75 & 57.11 \\

\midrule
\rowcolor{OpenBG}\multicolumn{4}{@{}c}{\textbf{Open-Weight MLLMs}}\\
\midrule

\makecell[{{p{4cm}}}]{GLM-4.5V} & 59.10 & 38.57 & 51.04 \\

\makecell[{{p{4cm}}}]{Llama 4 Maverick} & 56.74 & 36.83 & 55.39 \\

\makecell[{{p{4cm}}}]{Qwen3-VL-235B-A22B} & 72.29 & 38.35 & 50.83 \\

\makecell[{{p{4cm}}}]{Qwen3-Max} & 24.51 & 20.40 & 49.86 \\

\end{longtable}

\renewcommand{\arraystretch}{1.25}
\small
\setlength{\tabcolsep}{10pt}

\begin{longtable}{@{}p{4.5cm}ccccc@{}}
\caption{\centering Evaluation of large language models across five scientific benchmarks.}
\label{tab:five_benchmark_eval} \\
\toprule
\textbf{Model} & \textbf{ChemBench} & \textbf{ClimaQA} & \textbf{EarthSE} & \textbf{ProteinLM} & \textbf{MaScQA} \\
\midrule
\endfirsthead

\toprule
\textbf{Model} & \textbf{ChemBench} & \textbf{ClimaQA} & \textbf{EarthSE} & \textbf{ProteinLM} & \textbf{MaScQA} \\
\midrule
\endhead

\midrule
\endfoot

\bottomrule
\endlastfoot

\rowcolor{CloseBG}\multicolumn{6}{@{}c}{\textbf{Closed-Weight LLMs}}\\
\midrule

\makecell[{{p{4.5cm}}}]{Claude 4.5 Sonnet} & 69.20 & 71.27 & 66.60 & 15.57 & 83.23 \\

\makecell[{{p{4.5cm}}}]{Claude 4.1 Opus} & 69.02 & 71.94 & 71.80 & 11.65 & 83.23 \\

\makecell[{{p{4.5cm}}}]{GPT-5.1} & 63.90 & 76.45 & 86.44 & 68.22 & 61.08 \\

\makecell[{{p{4.5cm}}}]{GPT-5} & 70.42 & 81.62 & 80.20 & 59.22 & 93.54 \\

\makecell[{{p{4.5cm}}}]{GPT-4o} & 61.67 & 72.37 & 58.90 & 59.43 & 61.54 \\

\makecell[{{p{4.5cm}}}]{GPT-o3} & 69.81 & 81.77 & 81.40 & 66.10 & 93.85 \\

\makecell[{{p{4.5cm}}}]{Gemini-3-Pro} & 73.08 & 83.08 & 67.89 & 21.19 & 91.23 \\

\makecell[{{p{4.5cm}}}]{Gemini-2.5-Flash} & 42.32 & 75.12 & 72.30 & 4.34 & 62.15 \\

\makecell[{{p{4.5cm}}}]{Gemini-2.5-Pro} & 70.10 & 78.29 & 72.00 & 0.11 & 83.23 \\

\makecell[{{p{4.5cm}}}]{Grok-2-vision-1212} & 60.77 & 63.04 & 61.60 & 20.97 & 49.08 \\

\makecell[{{p{4.5cm}}}]{Seed1.6-vision} & 64.64 & 75.48 & 75.80 & 63.88 & 68.92 \\

\midrule
\rowcolor{OpenBG}\multicolumn{6}{@{}c}{\textbf{Open-Weight LLMs}}\\
\midrule

\makecell[{{p{4.5cm}}}]{Ling-flash-2.0} & 59.87 & 66.80 & 72.30 & 22.78 & 63.69 \\

\makecell[{{p{4.5cm}}}]{DeepSeek-R1} & 45.97 & 77.28 & 49.00 & 6.0381 & 44.46 \\

\makecell[{{p{4.5cm}}}]{GLM-4.5V} & 49.35 & 71.38 & 35.60 & 60.81 & 57.08 \\

\makecell[{{p{4.5cm}}}]{Kimi-k2} & 65.83 & 77.04 & 78.50 & 28.71 & 71.38 \\

\makecell[{{p{4.5cm}}}]{Llama 4 Maverick} & 66.44 & 68.06 & 56.00 & 19.70 & 80.77 \\

\makecell[{{p{4.5cm}}}]{Qwen3-VL-235B-A22B} & 63.68 & 77.39 & 75.20 & 50.11 & 81.85 \\

\makecell[{{p{4.5cm}}}]{Qwen3-Max} & 64.72 & 76.38 & 75.90 & 40.00 & 69.54 \\

\end{longtable}

\renewcommand{\arraystretch}{1.25}
\small
\setlength{\tabcolsep}{4pt}

\begin{longtable}{@{}p{3.5cm}cccccc@{}}
\caption{\centering Evaluation of large language models across six scientific benchmarks.}
\label{tab:textonly_eval} \\
\toprule
\textbf{Model} & \textbf{TRQA} & \textbf{CMPhysBench } & \textbf{PHYSICS} & \textbf{ResearchBench} & \textbf{SciCode } & \textbf{AstroVisBench}\\
\midrule
\endfirsthead

\toprule
\textbf{Model} & \textbf{TRQA} & \textbf{CMPhysBench } & \textbf{PHYSICS} & \textbf{ResearchBench} & \textbf{SciCode} & \textbf{AstroVisBench}\\
\midrule
\endhead

\midrule
\endfoot

\bottomrule
\endlastfoot

\rowcolor{CloseBG}\multicolumn{7}{@{}c}{\textbf{Closed-Weight LLMs}}\\
\midrule

\makecell[{{p{4.5cm}}}]{Claude 4.5 Sonnet} & 58.14 & 43.83 & 36.90 & 56.10 & 9.23 & 34.23 \\

\makecell[{{p{4.5cm}}}]{Claude 4.1 Opus} & 57.56 & 43.57 & 33.80 & 29.47 & 10.77 & 39.87 \\

\makecell[{{p{4.5cm}}}]{GPT-5.1} & 59.30 & 26.92 & 37.96 & 41.45 & 9.23 & 42.02 \\

\makecell[{{p{4.5cm}}}]{GPT-5} & 59.30 & 45.52 & 34.30 & 45.67 & 13.85 & 44.57 \\

\makecell[{{p{4.5cm}}}]{GPT-4o} & 51.16 & 35.37 & 28.80 & 33.04 & 4.62 & 30.73 \\

\makecell[{{p{4.5cm}}}]{GPT-o3} & 63.37 & 40.98 & 35.30 & 34.14 & 7.69 & 42.82 \\

\makecell[{{p{4.5cm}}}]{Gemini-3-Pro} & 59.88 & 55.97 & 34.40 & 61.51 & 16.92 & 42.23 \\

\makecell[{{p{4.5cm}}}]{Gemini-2.5-Flash} & 46.51 & 38.69 & 25.45 & 40.86 & 4.62 & 31.94 \\

\makecell[{{p{4.5cm}}}]{Gemini-2.5-Pro} & 52.33 & 38.17 & 31.75 & 50.73 & 6.15 & 43.40 \\

\makecell[{{p{4.5cm}}}]{Grok-2-vision-1212} & 45.35 & 28.27 & 28.15 & 49.63 & 7.69 & 33.51 \\

\makecell[{{p{4.5cm}}}]{Seed1.6-vision} & 45.93 & 43.18 & 35.30 & 45.00 & 10.77 & 32.21 \\

\midrule
\rowcolor{OpenBG}\multicolumn{7}{@{}c}{\textbf{Open-Weight LLMs}}\\
\midrule

\makecell[{{p{4.5cm}}}]{Ling-flash-2.0} & 34.88 & 46.35 & 29.60 & 50.29 & 3.08 & 48.12 \\

\makecell[{{p{4.5cm}}}]{DeepSeek-R1} & 48.26 & 52.10 & 33.50 & 49.73 & 0.00 & 0.11 \\

\makecell[{{p{4.5cm}}}]{GLM-4.5V} & 42.44 & 0.00 & 26.85 & 42.23 & 6.15 & 0.32 \\

\makecell[{{p{4.5cm}}}]{Kimi-k2} & 53.49 & 44.78 & 32.40 & 42.28 & 6.15 & 35.57 \\

\makecell[{{p{4.5cm}}}]{Llama 4 Maverick} & 52.33 & 44.29 & 33.65 & 38.31 & 4.62 & 31.91 \\

\makecell[{{p{4.5cm}}}]{Qwen3-VL-235B-A22B} & 47.67 & 64.75 & 35.10 & 40.62 & 6.15 & 29.84 \\

\makecell[{{p{4.5cm}}}]{Qwen3-Max} & 52.33 & 46.13 & 35.95 & 42.12 & 12.31 & 75.64 \\

\end{longtable}

To provide a more complete view of model capabilities, we present the full quantitative results of all evaluated models across both multimodal and text-only scientific tasks.

Table~\ref{tab:multimodal_eval} summarizes the performance of 19 representative models on three key multimodal scientific benchmarks: SLAKE, SFE, and MSEarth, covering visual reasoning on medical and scientific diagrams, code-based visualization, and entity localization.

Table~\ref{tab:textonly_eval} reports model-level performance on non-multimodal scientific reasoning tasks, such as causal inference, knowledge retrieval, mathematical derivation, and scientific QA, providing a comprehensive comparison of large language models across diverse text-only settings.

\section{Benchmark Description}
\label{app:bench_details}

\textbf{MSEarth.} MSEarth~\cite{zhao2025msearth} is a multimodal benchmark designed to assess scientific understanding in Earth science by integrating visual figures with textual reasoning. Its questions are drawn from curated figures and refined captions across atmosphere, cryosphere, hydrosphere, lithosphere, and biosphere, emphasizing not only perceptual recognition but also domain-informed inference. It serves as a representative task for evaluating multimodal scientific understanding and geophysical causal reasoning and reaches graduate-level.

\textbf{SLAKE.} SLAKE~\cite{liu2021slake} is a medical visual question‑answering benchmark designed for multimodal scientific reasoning in clinical imaging. Each instance is annotated with semantic segmentation masks and bounding‑boxes for key organs or structures, and is further structured around a built knowledge graph relational triples mapping organs, functions and disease entities. The dataset comprises real-world medical images drawn from three primary imaging modalities—CT, X-ray, and MRI—spanning anatomical regions such as the brain, neck, chest, abdomen, and pelvic cavity.

\textbf{SFE.} The SFE~\cite{zhou2025scientists} benchmark is a multimodal, multilingual evaluation suite designed to assess the scientific cognitive capacities of advanced models across perception, understanding, and reasoning. Drawing from authentic scientific raw data formats, SFE spans five high‑value disciplines (astronomy, chemistry, earth science, life science, and materials science) and comprises 66 expert‑curated tasks and 830 verified visual question‑answer pairs. The tasks are structured across three hierarchical cognitive levels—signal perception, attribute understanding, and comparative reasoning . By requiring models to process real scientific imagery and textual context, and to reason about them at an expert level, SFE pushes beyond superficial knowledge retrieval toward genuine scientific reasoning.

\textbf{AstroVisBench.} AstroVisBench~\cite{joseph2025astrovisbench} is a code‑centric benchmark designed to assess large language models’ capabilities in implementing scientific workflows and generating research‑quality visualizations within the domain of astronomy. Drawing from 110 publicly available Jupyter notebooks curated for astronomy research workflows, AstroVisBench require models to generate executable code for data processing and produce scientific visualizations conforming to domain standards.

\textbf{SciCode.} SciCode~\cite{tian2024scicode} is a scientist-curated benchmark that evaluates a model’s ability to translate natural-language research problems into executable Python solutions. Drawn from 80 challenging research tasks that span 16 sub-fields including mathematics, physics, chemistry, biology, and materials science, the benchmark decomposes each task into fine-grained sub-problems and each accompanied by gold-standard reference implementations and unit-test suites. SciCode emphasises realistic scientific workflows where problems often require domain-specific knowledge recall, multi-step reasoning, and calls to external scientific libraries.

\textbf{ChemBench.} ChemBench~\cite{Mirza2025} focuses on assessment of LLM's chemical knowledge comprehension and discipline-based reasoning across chemistry and materials science. Its subdomains ranges from general chemistry to more specialized fields such as inorganic, analytical and physical chemistry. Serving as a comprehensive probe of chemical intelligence,  ChemBench offers fine-grained insights into models' strength and weakness, making it an 

\textbf{ClimaQA.} ClimaQA~\cite{manivannan2024climaqa} transforms graduate-level climate science textbooks into scientifically grounded questions with domain-expert refinement. The benchmark contains two complementary subsets which are ClimaQA-Gold, manually curated and validated by experts, and ClimaQA-Silver, programmatically generated but aligned with the same scientific rigor, covering multiple QA formats such as multiple-choice, cloze-style, and free-form reasoning. By grounding tasks in authentic scientific content and emphasizing conceptual understanding, causal reasoning, and domain-specific inference, ClimaQA provides a more faithful lens to assess a model’s capability for climate science reasoning beyond general QA performance.

\textbf{EarthSE.} EarthSE~\cite{xu2025earthse} is developed to systematically probe LLMs’ competencies across the full breadth of Earth-science disciplines, covering five major spheres of Earth systems and 114 subfields, with diverse task formats tailored to evaluate foundational knowledge and domain-specific reasoning. Earth-Silver in particular is curated to represent professional-level difficulty, intended to test models’ depth of Earth-science knowledge and capability for scientific exploration. 

\textbf{ProteinLMBench.} ProteinLMBench~\cite{shen2024fine} aggregates a curated set of tasks drawn from widely used protein‐analysis datasets covering  protein-based property prediction, protein descriptions, and protein sequence
understanding and comprises 944 six-choice questions. Every item interleaves natural-language context with an amino-acid sequence span, forcing the model to align textual clues with residue patterns rather than rely on surface keyword cues. By sequence questions, ProteinLMBench offers a fine-grained, domain-specific complementary to broader biomolecular benchmarks.

\textbf{MaScQA.} MaScQA~\cite{zaki2024mascqa} is a specialized question-answering benchmark designed to evaluate large language models’ understanding and reasoning capabilities in materials science and metallurgical engineering. The questions are categorized into 14 domains: thermodynamics, atomic structure, mechanical behaviour, materials manufacturing, material applications, phase transition, electrical properties, material processing, transport phenomenon, magnetic properties, material characterization, fluid mechanics, material testing, and miscellaneous. This fine-grained categorization enables targeted evaluation of LLMs’ competence across diverse subfields, reflecting the interdisciplinary nature of modern materials science.

\textbf{TRQA.} TRQA~\cite{zhang2025origene} is a benchmark designed to evaluate large language models on biomedical reasoning and literature-based inference, designed to evaluate biomedical reasoning across literature evidence and real-world drug pipeline data. The benchmark targets a broad set of core capabilities: scientific planning, literature-grounded information retrieval, tool selection, reasoning toward biological conclusions, and critical self-evaluation.

\textbf{CMPhysBench.} CMPhysBench~\cite{wang2025cmphysbench} is designed to evaluate large language models’ scientific reasoning abilities in condensed matter physics, with a particular focus on symbolic derivation, algebraic manipulation and physical interpretation. Its questions span core subfields of condensed matter physics including magnetism, superconductivity, strongly correlated systems, etc. CMPhysBench serves as an evaluation to benchmark the depth of scientific reasoning in LLMs, particularly their ability to operate with formal representations, physical principles, and structured scientific logic.

\textbf{PHYSICS.} PHYSICS~\cite{zheng2025scaling} evaluates the ability of language models to perform undergraduate-level  physical reasoning spanning Mechanics, Electromagnetism, Thermodynamics, Optics, and Modern Physics. Each problem demands a combination of domain knowledge and symbolic reasoning ranging from conceptual understanding and detailed calculations. 

\textbf{ResearchBench.} ResearchBench~\cite{yu2024researchtown} aims to evaluate the capability of language models to simulate research workflows, particularly focusing on paper writing and peer-review generation. The benchmark consists of 1,000 paper-writing tasks and 200 review-writing tasks, all curated from recent top-tier machine learning venues such as NeurIPS 2024 and ICLR 2024. This benchmark assesses a model's ability to generate research ideas, understand technical contributions and provide coherent scientific writing aligned with the realistic academic standards.

\section{Representative Task Cases}

\subsection{MaScQA}

\begin{tcolorbox}[
    colback = yellow!5!white, 
    colframe = yellow!50!black, 
    colbacktitle = yellow!75!black,
    title = MaScQA — Thermodynamic Reasoning in Materials Science,
    breakable
]
\textbf{Question:} \\
A student aims to deposit a thin metallic film on $SiO_2$ substrate, with an adhesion layer between the metal film and substrate, in a contiguous planar fashion. Island type of growth must be avoided. The student performs an extensive optimization exercise. Which one of the following steps is in the right direction?   (A) Choose a metallic adhesion layer with very low interfacial energy with the deposited thin film (B) Choose a metallic adhesion layer with very low interfacial energy with $SiO_2$, irrespective of its interaction with metal film to be deposited (C) Increase the substrate temperature and decrease the deposition rate (D) Use intermittent stages of deposition followed by annealing\\

\textbf{Question Type:} Multi-Choice Question \\

\textbf{Answer:} {\color{red}{A}}
\tcblower

\textbf{Response from Gemini 3 Pro:} \\
To achieve the growth of a contiguous planar thin metallic film and avoid island (Volmer-Weber) growth, one must ensure that the deposited film "wets" the surface beneath it.

**Theoretical Analysis:**
The growth mode is determined by the balance of surface and interfacial energies, described by Young's equation and the spreading parameter $S$. For a film ($f$) depositing on a substrate ($s$), the condition for layer-by-layer (planar) growth is:
$$ \gamma_s \ge \gamma_f + \gamma_{int} $$
where:
*   $\gamma_s$ is the surface energy of the substrate (in this case, the adhesion layer).
*   $\gamma_f$ is the surface energy of the depositing metal film.
*   $\gamma_{int}$ is the interfacial energy between the film and the substrate.

**Evaluating the Options:**

*   **(A) Choose a metallic adhesion layer with very low interfacial energy with the deposited thin film:**
    In this scenario, the "substrate" for the top metal film is the adhesion layer. To satisfy the wetting condition ($\gamma_{AdhesionLayer} \ge \gamma_{MetalFilm} + \gamma_{Interface}$), it is crucial to minimize $\gamma_{Interface}$. A very low interfacial energy implies strong bonding (affinity) between the adhesion layer and the metal film, which promotes wetting and planar growth. This is the correct direction.

*   **(B) Choose a metallic adhesion layer with very low interfacial energy with $SiO_2$, irrespective of its interaction with metal film to be deposited:**
    While the adhesion layer must adhere to the $SiO_2$, ignoring the interaction with the top metal film is incorrect. If the interfacial energy between the adhesion layer and the top metal film is high (poor bonding), the inequality $\gamma_{AdhesionLayer} \ge \gamma_{MetalFilm} + \gamma_{Interface}$ will likely not be met, leading to island growth of the metal film on top of the adhesion layer.

*   **(C) Increase the substrate temperature and decrease the deposition rate:**
    High substrate temperatures increase surface diffusivity. In systems that are not perfectly wetting (which is common for metals on dissimilar surfaces), high mobility allows atoms to aggregate into thermodynamically stable 3D islands (agglomeration). To force a continuous film in a system that might tend toward islanding, one typically uses *lower* substrate temperatures and *higher* deposition rates to increase nucleation density and freeze the atoms in a continuous layer (kinetic control). Increasing temperature generally promotes islanding and dewetting for thin metal films.

*   **(D) Use intermittent stages of deposition followed by annealing:**
    Annealing provides thermal energy which drives the system toward thermodynamic equilibrium. For thin metallic films, the equilibrium state is often a dewetted set of droplets (islands) to minimize total surface energy (Rayleigh instability). Therefore, annealing usually destroys the continuity of a thin film rather than promoting it.

**Conclusion:**
Only option (A) correctly addresses the thermodynamic requirements to prevent island growth and promote a contiguous planar film.

\textbf{\color{red}{[A]}} \\

\textbf{Judgment:} {\color{red}{Correct.}} \\

\textbf{Note}: Gemini 3 Pro is the only model that answers this question correctly.
\end{tcolorbox}

\subsection{Chembench}

\begin{tcolorbox}[
    colback = yellow!5!white, 
    colframe = yellow!50!black, 
    colbacktitle = yellow!75!black,
    title = Chembench — Analytical Chemistry (Molecular Vibrations),
    breakable
]
\textbf{Question:} \\
What is the total number of normal vibrational modes for a $\text{C}_{60}$ molecule? \\
(A) 174 \quad (B) 180 \quad (C) 54 \quad (D) 60 \\

\textbf{Question Type:} Multi-Choice Question \\

\textbf{Answer:} {\color{red}{A}}
\tcblower

\textbf{Response from Gemini 3 Pro:} \\
To determine the total number of normal vibrational modes for a $\text{C}_{60}$ molecule, we follow these steps:

1.  \textbf{Identify the number of atoms ($N$):} The molecule is $\text{C}_{60}$, which consists of 60 carbon atoms. So, $N = 60$.
2.  \textbf{Calculate the total degrees of freedom:} The total degrees of freedom for a molecule with $N$ atoms is given by $3N$.
    $$3 \times 60 = 180$$
3.  \textbf{Subtract translational and rotational degrees of freedom:}
    \begin{itemize}
        \item There are always 3 translational degrees of freedom.
        \item Since $\text{C}_{60}$ is a non-linear molecule (it has a spherical cage-like structure), it has 3 rotational degrees of freedom.
        \item The formula for the number of vibrational modes for a non-linear molecule is $3N - 6$.
    \end{itemize}
4.  \textbf{Perform the calculation:}
    $$ \text{Vibrational modes} = 3(60) - 6 = 180 - 6 = 174 $$

Therefore, the total number of normal vibrational modes is 174.

\textbf{\color{red}{[ANSWER]A[/ANSWER]}} \\

\textbf{Judgment:} {\color{red}{Correct.}} \\

\textbf{Note}: Gemini 3 Pro accurately applies the degrees of freedom formula ($3N-6$) for non-linear molecules to solve this structural chemistry problem.
\end{tcolorbox}

\begin{tcolorbox}[
    colback = yellow!5!white, 
    colframe = yellow!50!black, 
    colbacktitle = yellow!75!black,
    title = Chembench — Analytical Chemistry (NMR Spectroscopy),
    breakable
]
\textbf{Question:} \\
Determine the number of peaks in the 1H NMR spectrum of the molecule with the SMILES \texttt{OCC1=CNC(=O)C1=O}. \\

\textbf{Question Type:} Numeric Question \\

\textbf{Answer:} {\color{red}{4}}
\tcblower

\textbf{Response from Gemini 3 Pro:} \\
To determine the number of peaks in the 1H NMR spectrum of the molecule with the SMILES \texttt{OCC1=CNC(=O)C1=O}, we first analyze its structure.

1.  \textbf{SMILES Parsing:}
    \begin{itemize}
        \item \texttt{OCC}: A hydroxymethyl group (\texttt{HO-CH2-}) attached to the ring anchor.
        \item \textbf{Ring Structure:} The connectivity corresponds to a \textbf{1H-pyrrole-2,3-dione} core substituted at position 4.
        \item \textbf{Substituent:} A hydroxymethyl group (\texttt{-CH2OH}) at position 4.
    \end{itemize}

2.  \textbf{Identify Proton Environments:}
    \begin{itemize}
        \item \textbf{Hydroxyl Proton (-OH):} The proton on the oxygen of the hydroxymethyl group. This represents \textbf{1} distinct chemical environment.
        \item \textbf{Methylene Protons (-CH2-):} The two protons on the carbon of the hydroxymethyl group. Since the molecule has no chiral center and the ring is planar, these two protons are enantiotopic (equivalent). They represent \textbf{1} distinct chemical environment.
        \item \textbf{Vinylic Proton (=CH-):} The proton on the ring carbon at position 5 (adjacent to the nitrogen). This represents \textbf{1} distinct chemical environment.
        \item \textbf{Amide/Imide Proton (-NH-):} The proton on the ring nitrogen. This represents \textbf{1} distinct chemical environment.
    \end{itemize}

3.  \textbf{Count the Peaks:}
    There are 4 distinct sets of equivalent protons: (1) -OH, (2) -CH2-, (3) =CH-, (4) -NH-. Assuming a standard solvent where exchangeable protons are visible, we expect 4 signals.

\textbf{\color{red}{[ANSWER]4[/ANSWER]}} \\

\textbf{Judgment:} {\color{red}{Correct.}} \\

\textbf{Note}: Gemini 3 Pro correctly parses the SMILES string, identifies equivalent protons based on molecular symmetry, and provides the exact integer count required.
\end{tcolorbox}






\subsection{SciCode}

\begin{tcolorbox}[
    colback = yellow!5!white,
    colframe = yellow!50!black,
    colbacktitle = yellow!75!black,
    title = SciCode,
    breakable
]

\textbf{Question:} \\
Create a function to solve the linear system $\mathbf{A} \mathbf{x} = \mathbf{b}$ using the conjugate gradient method. This function takes a matrix $\mathbf{A}$ and a vector $\mathbf{b}$ as inputs.

\textbf{Answer:}

\begin{lstlisting}[language=Python]
def cg(A, b, x, tol):
    """
    Inputs:
    A : Matrix, 2d array size M * M
    b : Vector, 1d array size M
    x : Initial guess vector, 1d array size M
    tol : tolerance, float
    Outputs:
    x : solution vector, 1d array size M
    """
    # Initialize residual vector
    res = b - np.dot(A, x)
    # Initialize search direction vector
    search_direction = res.copy()
    # Compute initial squared residual norm
    old_res_norm = np.linalg.norm(res)
    itern = 0
    # Iterate until convergence
    while old_res_norm > tol:
        A_search_direction = np.dot(A, search_direction)
        step_size = old_res_norm**2 / np.dot(search_direction, A_search_direction)

        # Update solution
        x += step_size * search_direction
        # Update residual
        res -= step_size * A_search_direction
        new_res_norm = np.linalg.norm(res)

        # Update search direction vector
        search_direction = res + (new_res_norm / old_res_norm)**2 * search_direction

        # Update squared residual norm for next iteration
        old_res_norm = new_res_norm
        itern += 1
    return x
\end{lstlisting}

\end{tcolorbox}

\subsection{PHYSICS}

\begin{tcolorbox}[
    colback = yellow!5!white,
    colframe = yellow!50!black,
    colbacktitle = yellow!75!black,
    title = PHYSICS,
    breakable
]

\textbf{Question:} \\
To make a flat stone skip across the water surface when thrown quickly, the stone may bounce and fly towards the distance, commonly known as "stone skipping." To achieve the "stone skipping" effect, the angle between the direction of the stone's velocity and the water surface at the point of contact must not exceed $\theta$. To observe "stone skipping," a student throws a stone horizontally from a height $h$ above the water surface. What is the minimum launch velocity required? (Neglect air resistance during the stone's flight, and the acceleration due to gravity is $g$.)

\textbf{Answer:} { \color{red}{
${\frac{\sqrt{2 g h}}{\tan \theta}}$
}

}

\end{tcolorbox}

\subsection{CMPhysBench}

\begin{tcolorbox}[
    colback = yellow!5!white,
    colframe = yellow!50!black,
    colbacktitle = yellow!75!black,
    title = CMPhysBench - Theoretical Foundations,
    breakable
]

\textbf{Question:} \\
A particle of mass $m$ is in the ground state of a one-dimensional harmonic oscillator potential

\begin{equation*}
V_{1}(x)=\frac{1}{2} k x^{2}, \quad k>0
\end{equation*}

When the spring constant $k$ suddenly changes to $2k$, the potential then becomes

\begin{equation*}
V_{2}(x)=k x^{2}
\end{equation*}

Immediately measure the energy of the particle, and find the expression for the probability of the particle being in the ground state of the new potential $V_{2}$.

\textbf{Answer:}

(a) The wave function of the particle $\psi(x, t)$ should satisfy the time-dependent Schrödinger equation

\begin{equation*}
\mathrm{i} \hbar \frac{\partial}{\partial t} \psi=-\frac{\hbar^{2}}{2 m} \frac{\partial^{2}}{\partial x^{2}} \psi+V \psi \tag{3}
\end{equation*}

When $V$ undergoes a sudden change (from $V_{1} \rightarrow V_{2}$) but with a finite change quantity, $\psi$ remains a continuous function of $t$, implying that $\psi$ does not change when $V$ changes abruptly.

Denote $\psi_{0}(x)$ and $\phi_{0}(x)$ as the ground state wave functions of the potential $V_{1}$ and $V_{2}$, respectively. After the potential suddenly changes from $V_{1}$ to $V_{2}$, the wave function of the particle remains $\psi_{0}$. The probability of measuring the particle in the state $\phi_{0}$ is $|\langle\psi_{0} \mid \phi_{0}\rangle|^{2}$.

Rewrite $V_{1}$ and $V_{2}$ in standard form:

\begin{gather*}
V_{1}(x)=\frac{1}{2} k x^{2}=\frac{1}{2} m \omega_{1}^{2} x^{2} \tag{$\prime$}\\
V_{2}(x)=k x^{2}=\frac{1}{2} m \omega_{2}^{2} x^{2} \tag{$\prime$}
\end{gather*}

It is clear that

\begin{equation*}
\omega_{2}=\sqrt{2} \omega_{1} \tag{4}
\end{equation*}

$\psi_{0}$ and $\phi_{0}$ can be expressed as in formulas (3) and (5) from problem 3.2, namely

\begin{align}
\psi_{0}(x) &= \left( \frac{\alpha}{\sqrt{\pi}} \right)^{\frac{1}{2}} \mathrm{e}^{-\alpha^2 x^2 / 2}, &
\alpha^{2} &= m \omega_{1} / \hbar \notag \\
\phi_{0}(x) &= \left( \frac{\beta}{\sqrt{\pi}} \right)^{\frac{1}{2}} \mathrm{e}^{-\beta^2 x^2 / 2}, &
\beta^{2} &= m \omega_{2} / \hbar \tag{6}
\end{align}

where

\begin{equation*}
\beta^{2} / \alpha^{2}=\omega_{2} / \omega_{1}=\sqrt{2} \tag{7}
\end{equation*}

Thus

\begin{align*}
\langle\psi_{0} \mid \phi_{0}\rangle & =\sqrt{\frac{\alpha \beta}{\pi}} \int_{-\infty}^{+\infty} \mathrm{e}^{-\frac{1}{2}(\alpha^{2}+\beta^{2}) x^{2}} \mathrm{~d} x=(\frac{2 \alpha \beta}{\alpha^{2}+\beta^{2}})^{\frac{1}{2}} \\
|\langle\psi_{0} \mid \phi_{0}\rangle|^{2} & =\frac{2 \alpha \beta}{\alpha^{2}+\beta^{2}}=\frac{2 \beta / \alpha}{1+\beta^{2} / \alpha^{2}}=\frac{2^{5 / 4}}{1+\sqrt{2}}=0.9852 \tag{8}
\end{align*}

This is the required probability.
(b) Consider the time when the potential changes for the first time $(V_{1} \rightarrow V_{2})$ as $t=0$, then the wave function is

\begin{equation*}
\psi(x, 0)=\psi_{0}(x) \tag{9}
\end{equation*}

Let $\phi_{n}(x)$ denote the energy eigenstates of the potential $V_{2}$, corresponding to energy levels

$E_{n}=(n+\frac{1}{2}) \hbar \omega_{2}$

Expand $\psi_{0}$ as a linear combination of $\phi_{n}$,

\begin{equation*}
\psi_{0}(x)=\sum_{n} C_{n} \phi_{n}(x), \quad(n \text { can only take even values }) \tag{10}
\end{equation*}

For $0<t<\tau$, in Schrödinger equation (3) $V=V_{2}(x)$, its solution is

\begin{align*}
\psi(x, t) & =\sum_{n} C_{n} \phi_{n}(x) \mathrm{e}^{-\mathrm{i} E_{n} / / \hbar} \\
& =\mathrm{e}^{-i \omega_{2} t \cdot 2} \sum_{n} C_{n} \phi_{n}(x) \mathrm{e}^{-i \omega_{\omega_{2}} t} \tag{11}
\end{align*}

To have $\psi(x, \tau)=A \psi_{0}(x)$, it must hold that

\begin{equation*}
\mathrm{e}^{-\mathrm{in} \omega_{2} \tau}=1, \quad n=0,2,4, \cdots \tag{12}
\end{equation*}

That is

\begin{equation*}
\mathrm{e}^{\mathrm{i} \omega_{2} \tau}= \pm 1 \tag{12'}
\end{equation*}

The $\tau$ that satisfies this condition is

\begin{equation*}
\tau=l \pi / \omega_{2}=l \pi \sqrt{\frac{m}{2 k}}, \quad l=1,2,3, \cdots \tag{13}
\end{equation*}

When $t=\tau$, after the potential changes from $V_{2}$ back to $V_{1}$, the particle remains in the state $\psi_{0}$, with energy $E=\hbar \omega_{1} / 2$.

\end{tcolorbox}

\subsection{ClimaQA}

\begin{tcolorbox}[
    colback = yellow!5!white, 
    colframe = yellow!50!black, 
    colbacktitle = yellow!75!black,
    title = ClimaQA,
    breakable
]
\textbf{Question:} \\
Why does the reflectivity increase between points 3 and 4 in the vertical profile of radar data for stratiform precipitation?

A) Due to the reduction in relative humidity below cloud, causing evaporation of raindrops.

B) As a result of the snowflakes becoming more compact and collapsing due to increased terminal velocity.

C) Due to the reduction in size of snowflakes as they melt and acquire a liquid layer.

D) Because of the increase in the concentration of ice crystals in that layer.

\textbf{Question Type:} Multi-Choice Question \\

\textbf{Answer:} {\color{red}{C}}

\end{tcolorbox}

\subsection{EarthSE}

\begin{tcolorbox}[
    colback = yellow!5!white, 
    colframe = yellow!50!black, 
    colbacktitle = yellow!75!black,
    title = EarthSE,
    breakable
]
\textbf{Question:} \\
	
How do varying fire regimes and browsing intensities influence vegetation dynamics and species composition in Mediterranean ecosystems, particularly in scenarios modeled near Lago di Massacciucoli during the Holocene, and what implications do interacting disturbances have on drought-sensitive species such as Abies alba and the dominance of Quercus ilex under different climatic conditions? \\

\textbf{Question Type:} Analysis \\

\textbf{Answer:} 

Varying fire regimes and browsing intensities serve as critical factors in shaping vegetation dynamics and species composition in Mediterranean ecosystems. In simulations near Lago di Massacciucoli, small, infrequent fires coupled with low browsing intensity promote the coexistence of Abies alba and Quercus ilex in mixed forests. However, large, infrequent fires decrease overall biomass and favor shade-tolerant species such as evergreen oaks and deciduous hardwoods. High fire frequency paired with warmer-than-present summer temperatures significantly reduces A. alba biomass, suggesting its vulnerability to such disturbances. Browsing intensities further exacerbate this effect, with moderate browsing eliminating A. alba regeneration entirely, favoring species like Q. ilex and shade-intolerant evergreen and deciduous competitors. Interactions between fire, browsing, and climate highlight complex dynamics, where increased light availability and reduced competition from A. alba under higher disturbance levels allow Q. ilex to dominate. These findings emphasize the combined impact of anthropogenic activity and climatic changes on biodiversity and forest compositional shifts in Mediterranean regions.

\end{tcolorbox}

\subsection{ProteinLMBench}

\begin{tcolorbox}[
    colback = yellow!5!white, 
    colframe = yellow!50!black, 
    colbacktitle = yellow!75!black,
    title = ProteinLMBench,
    breakable
]
\textbf{Question:} \\
	
Analyze the sequence of the following enzyme protein, and step by step deduce which reaction(s) this enzyme is involved in? \\
<seq> M S I P S S Q Y G F V F N K Q S G L K L R N D L P V H K P K A G Q L L L K V D A V G L C H S D L H V I Y E G L D C G D N Y V M G H E I A G T V A A V G D D V I N Y K V G D R V A C V G P N G C G G C K Y C R G A I D N V C K N A F G D W F G L G Y D G G Y Q Q Y L L V T R P R N L S R I P D N V S A D V A A A S T D A V L T P Y H A I K M A Q V S P T S N I L L I G A G G L G G N A I Q V A K A F G A K V T V L D K K K E A R D Q A K K L G A D A V Y E T L P E S I S P G S F S A C F D F V S V Q A T F D V C Q K Y V E P K G V I M P V G L G A P N L S F N L G D L A L R E I R I L G S F W G T T N D L D D V L K L V S E G K V K P V V R S A K L K E L P E Y I E K L R N N A Y E G R V V F N P </seq> \\

\textbf{Question Type:} Analysis \\

\textbf{Answer:} 

Based on the analysis of the provided protein sequence, the function of this protein is described as follows: A zinc protein. Acts on primary or secondary alcohols or hemi-acetals with very broad specificity; however the enzyme oxidizes methanol much more poorly than ethanol. The animal, but not the yeast, enzyme acts also on cyclic secondary alcohols. So, this enzyme should belong to the alcohol:NAD<sup>+</sup> oxidoreductase. Therefore, by synthesizing the analysis of this enzyme, we can determine the reaction(s) it is involved in as follows:
(1) a primary alcohol + NAD<sup>+</sup> = an aldehyde + NADH + H<sup>+</sup> (2) a secondary alcohol + NAD<sup>+</sup> = a ketone + NADH + H<sup>+</sup> .

\end{tcolorbox}

\subsection{TRQA}

\begin{tcolorbox}[
    colback = yellow!5!white, 
    colframe = yellow!50!black, 
    colbacktitle = yellow!75!black,
    title = TRQA,
    breakable
]
\textbf{Question:} \\
What is the highest global development stage achieved by drugs targeting PDE2A for cognitive impairment in schizophrenia up to April 2025, and which drugs are in this stage?


\textbf{Answer:} {\color{red}{Phase I; BI 474121}}
\end{tcolorbox}

\subsection{ResearchBench}

\begin{tcolorbox}[
    colback = yellow!5!white, 
    colframe = yellow!50!black, 
    colbacktitle = yellow!75!black,
    title = ResearchBench,
    breakable
]
\textbf{Question:} \\
You are helping with the scientific hypotheses generation process. We in general split the period of conducting research into four steps. Firstly it's about finding a good and specific background research question, and an introduction of the previous methods under the same topic; Secondly its about finding inspirations (mostly from literatures), which combined with the background research question, can lead to a impactful research hypothesis; Thirdly it's hypothesis generation based on the background research question and found inspirations; Finally it's about designing and conducting experiments to verify hypothesis. An example is the backpropagation of neural networks. In backpropagation, the research question is how to use data to automatically improve the parameters of a multi-layer logistic regression, the inspiration is the chain rule in mathematics, and the research hypothesis is the backpropagation itself. In their paper, the authors have conducted experiments to verify their hypothesis. Now we have identified a good research question, and we have found a core inspiration in a literature for this research question. But one inspiration might not have enough information to support a finding of novel, valid, and significant research hypothesis. Therefore we also have found a series of inspiration candidates, which might provide additional useful information to assist the core inspiration for the next step of hypothesis generation. Please help us generate a novel, valid, and significant research hypothesis based on the background research question and the inspirations. 
The background research question is: What are the processes and sources responsible for the formation and evolution of dust in the early universe, particularly at redshifts greater than 4?

The introduction of the previous methods is:Survey not provided. Please overlook the survey.

The core inspiration is: Title: Dust formation in primordial Type II supernovae; Abstract: We have investigated the formation of dust in the ejecta of Type II supernovae (SNe), mostly of primordial composition, to answer the question of where are the first solid particles formed in the universe. However, we have also considered non-zero progenitor’s metallicity values up to Z = Z$\odot$. The calculations are based on standard nucleation theory and the scheme has been first tested on the well studied case of SN1987A, yielding results that are in agreement with the available data. We find that: i) the first dust grains are predominantly made of silicates, amorphous carbon (AC), magnetite, and corundum; ii) the largest grains are the AC ones, with sizes around 300u , whereas other grain types have smaller radii, around 10-20u The grain size distribution depends somewhat on the thermodynamics of the ejecta expansion and variations in the results by a factor   2 might occur within reasonable estimates of the relevant parameters. Also, and for the same reason, the grain size distribution, is essentially unaffected by metallicity changes. The predictions on the amount of dust formed are very robust: for Z = 0, we find that SNe with masses in the range (12-35)M$\odot$ produce about 0.08M$\odot$ < Md <   0.3M$\odot$ of dust/SN. The above range increases by roughly 3 times as the metallicity is increased to solar values. We discuss the implications and the cosmological consequences of the results..

The additional inspiration candidates are: - Composition and quantities of dust produced by AGB-stars and returned to the interstellar medium
- The evolution of refractory interstellar grains in the solar neighborhood

Now you have seen the background research question, the core inspiration, and many potential additional inspiration candidates. Please try to generate a novel, valid, and significant research hypothesis based on the background research question and the inspirations. (response format: 'Hypothesis: 
Reasoning Process:
')

\textbf{Answer:} {\color{red}{The authors propose that multiple stellar sources and processes, particularly supernovae and asymptotic giant branch (AGB) stars, significantly contribute to the formation and evolution of dust in the early universe. They suggest that the combination of these sources, along with dust growth in the interstellar medium, can account for the observed dust masses in early galaxies.}}
\end{tcolorbox}

\subsection{MSEarth}

\begin{tcolorbox}[
    colback = yellow!5!white, 
    colframe = yellow!50!black, 
    colbacktitle = yellow!75!black,
    title = MSEarth,
    breakable
]
\textbf{Question:} \\
Delineation of hazardous regions for the nine classiﬁcations for   $P(\mathfrak{u};z_{p}|(\mathfrak{n}))\geq0.5$  Fe–Mn is the Fe–Mn hazard; N is the nitrogen hazard; As is the As hazard; saline is the saline hazard.

Which aquifer shows the highest spread of the Fe-Mn hazard?

{\centering
\includegraphics[width=0.4\linewidth]{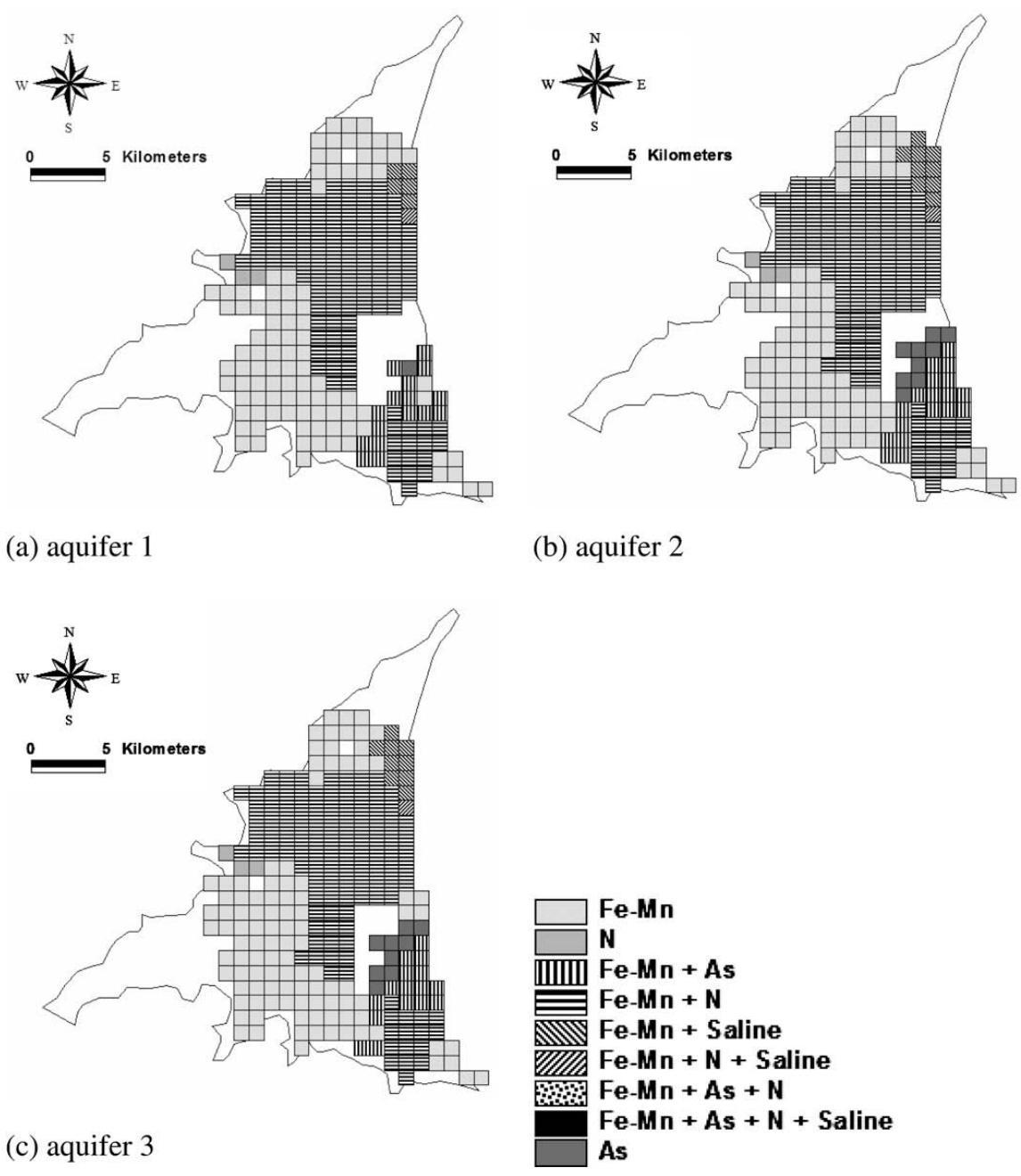}
\par}

(A) Aquifer 1 \quad (B) Aquifer 2 \quad (C) Aquifer 3 \quad (D) None of the above \\

\textbf{Question Type:} Multi-Choice Question \\

\textbf{Answer:} {\color{red}{C}}
\end{tcolorbox}

\subsection{AstroVisBench}

\begin{tcolorbox}[
    colback = yellow!5!white,
    colframe = yellow!50!black,
    colbacktitle = yellow!75!black,
    title = AstroVisBench — Data Processing and Visualization in Astrophysics,
    breakable
]
\textbf{Question:} \\
\textbf{Setup:} To set up the environment for our analysis, import the necessary astronomical and data handling packages that will allow us to perform sky coordinate calculations, access and handle data tables, and make HTTP requests to retrieve data. Additionally, configure the environment to properly display tables and set up appropriate input and output stream handling capabilities for image processing. Make sure that each imported module serves its purpose, whether that's astrometric calculations, pretty printing, image handling or data visualization. \\

\textbf{Processing:} For this analysis, utilize a series of functions to query an astronomical database using an API. These functions should enable you to perform a precise HSC catalog search using positional queries with an angular search radius, while also being able to retrieve metadata about the tables and columns available in the database. Implement the necessary error checking to ensure the compatibility of request parameters, and validate that query format and table specification adhere to the accepted values. Then, use the astropy name resolver to acquire the celestial coordinates of a specified astronomical target and filter the data to select only those entries that meet certain criteria, such as having a minimum number of measurements in specified bands. \\

\textbf{Visualization:} Plot the celestial coordinates of the extracted objects on a 2D visualization to represent their positions in relation to the center of the galaxy. Use distinct markers to denote both the location of the galaxy center and the objects obtained from the catalog search, ensuring that their sizes and colors are visibly distinct to differentiate the two. Set appropriate axes labels for Right Ascension and Declination, with an inverted x-axis to align with conventional celestial plotting practices. Add a legend to facilitate the understanding of the symbols used in the plot, and ensure aspect ratio maintenance for accurate representation. \\

\textbf{Question Type:} Code Generation (Multi-step: Processing \& Visualization) \\

\textbf{Answer:} {\color{red}{Code}}
\tcblower

\textbf{Response from Intern-S1:} \\

\textbf{Processing Code:}
\begin{lstlisting}[language=Python]
def get_target_coordinates(target_name):
    try:
        coord = SkyCoord.from_name(target_name)
        return coord.ra.deg, coord.dec.deg
    except Exception as e:
        print(f"Error resolving target name: {e}")
        raise

def query_hsc_catalog(ra, dec, radius, constraints):
    """Query the HSC catalog with a cone search and constraints."""
    if radius > 2.0:
        raise ValueError("Radius must be <= 2 degrees for HSC cone search.")
    
    base_url = "https://catalogs.mast.stsci.edu/api/v0.1/hsc/cone.csv"
    params = {
        'ra': ra,
        'dec': dec,
        'radius': radius,
        'where': constraints,
        'format': 'csv'
    }
    
    response = requests.get(base_url, params=params)
    if response.status_code == 200:
        table = Table.read(BytesIO(response.content), format='csv')
        return table
    else:
        print(f"Error in API request: {response.status_code}")
        print(response.text)
        response.raise_for_status()

def get_hsc_metadata():
    """Retrieve metadata about HSC tables and columns."""
    metadata_url = "https://catalogs.mast.stsci.edu/api/v0.1/hsc/metadata"
    response = requests.get(metadata_url)
    if response.status_code == 200:
        return response.json()
    else:
        print(f"Error retrieving metadata: {response.status_code}")
        response.raise_for_status()

# Example usage for IC 1613 with 0.5 degree radius and measurement constraints
target_name = "IC 1613"
radius_deg = 0.5
constraints = "A_F475W_N >= 10 AND A_F814W_N >= 10"

try:
    ra, dec = get_target_coordinates(target_name)
    print(f"Coordinates for {target_name}: RA={ra:.4f}, Dec={dec:.4f}")
    
    result_table = query_hsc_catalog(ra, dec, radius_deg, constraints)
    print(f"\nRetrieved {len(result_table)} entries matching the criteria.")
    print("First 5 entries:")
    result_table[:5].pprint(max_width=150)
    
except Exception as e:
    print(f"An error occurred: {e}")
\end{lstlisting}

\textbf{Processing Execution Output (API Error):}
\begin{verbatim}
Coordinates for IC 1613: RA=16.2017, Dec=2.1195
Error in API request: 404
{"detail":"Not Found"}
An error occurred: 404 Client Error: Not Found for url: ...
\end{verbatim}

\textbf{Visualization Code (Assuming data table 'tab' is available):}
\begin{lstlisting}
# Visualization

# Create a figure and axis
plt.figure(figsize=(10, 8))

# Plot the catalog objects as blue points
plt.scatter(tab['MatchRA'], tab['MatchDec'], s=20, c='skyblue', edgecolor='k', alpha=0.6, label='HSC Objects')

# Plot the galaxy center as a large red star
plt.scatter(ra_ic1613, dec_ic1613, s=200, c='red', marker='*', label='IC 1613 Center')

# Invert the x-axis to align with celestial plotting conventions
plt.gca().invert_xaxis()

# Add labels, title, and legend
plt.xlabel('Right Ascension (degrees)')
plt.ylabel('Declination (degrees)')
plt.title('Spatial Distribution of Objects in IC 1613 Field')
plt.legend()
plt.grid(True)

# Display the plot
plt.show()
\end{lstlisting}

\textbf{Visualization Result:}

\includegraphics[width=\linewidth]{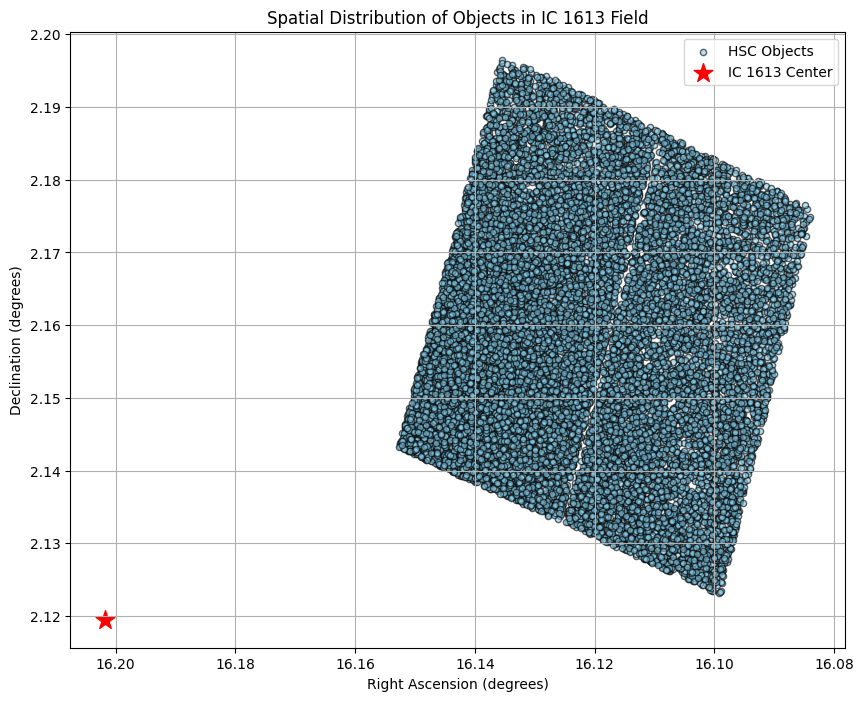}

\textbf{Judgment Rationale for Visualization (V2):} \\
Comparing the under-test image with the ground truth image, several styling differences are observed: 1) The marker style for HSC objects differs (filled circles vs. simple dots). 2) The marker size differs (s=20 vs. markersize=1). 3) The galaxy center marker differs (filled red star vs. red 'x' marker). 4) The model includes a grid and title which are absent in the ground truth. Despite these aesthetic differences, both visualizations convey the same key information: the spatial distribution of HSC measurements in relation to the IC 1613 galaxy center, with an inverted x-axis, proper axis labels, distinct markers for objects vs. center, and a legend. The overall pattern, data positions, and coordinate ranges are identical. The differences are primarily aesthetic rather than informational. \\

\textbf{Overall Judgment (Processing \& Visualization):} \\
Processing Code is functionally correct but execution fails due to external API error (404 Not Found). \\
Visualization Code executes successfully but has minor aesthetic differences compared to the target visualization. \\
\textbf{Visualization Error Type:} {\color{red}{Minor Error}}
\end{tcolorbox}

\subsection{SLAKE}

\begin{tcolorbox}[
    colback = yellow!5!white, 
    colframe = yellow!50!black, 
    colbacktitle = yellow!75!black,
    title = SLAKE,
    breakable
]

\textbf{Question:} \\
What modality is used to take this image?

\textbf{Image:} \\

{\centering
\includegraphics[width=0.4\linewidth]{imgs/appendix/SLAKE_0.png}
\par}


\textbf{Answer:} {\color{red}{CT}}

\end{tcolorbox}

\subsection{SFE}

\begin{tcolorbox}[
    colback = yellow!5!white, 
    colframe = yellow!50!black, 
    colbacktitle = yellow!75!black,
    title = SFE,
    breakable
]
\textbf{Question:} \\
Given an electronic band structure plot, analyze the positions of the valence band maximum (VBM) and conduction band minimum (CBM) to decide if the material has a direct bandgap (VBM and CBM at the same high‑symmetry k‑point, which you must specify), an indirect bandgap (VBM and CBM at different high‑symmetry k‑points, which you must specify both), or if band crossings at the Fermi level indicate a metallic state with no bandgap. Reply with a single concise narrative paragraph stating your conclusion, including the relevant high‑symmetry k‑points for direct or indirect cases, or for a metallic case briefly explaining that band crossings preclude any bandgap, without bullet points, lists, or additional commentary. 

Based on the provided band structure <image>, determine whether the material exhibits a direct or indirect bandgap.

{\centering
\includegraphics[width=0.4\linewidth]{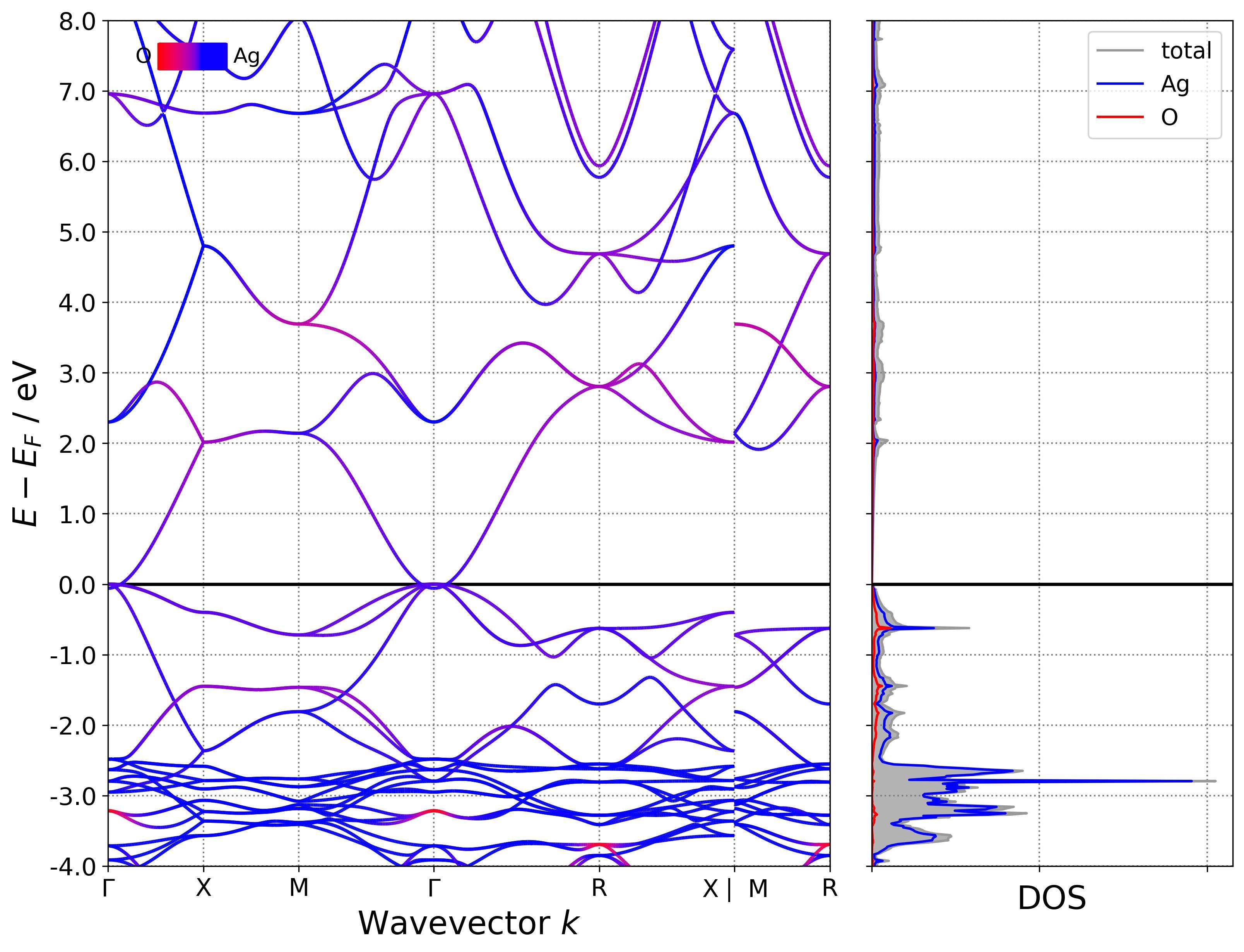}
\par}


\textbf{Answer:} {\color{red}{The material is metallic, doesn't exhibit a direct or indirect bandgap.}}
\end{tcolorbox}

\end{document}